\documentclass[pdflatex,sn-mathphys-ay]{sn-jnl}

\usepackage{graphicx}%
\usepackage{multirow}%
\usepackage{amsmath,amssymb,amsfonts}%
\usepackage{amsthm}%
\usepackage{mathrsfs}%
\usepackage[title]{appendix}%
\usepackage{xcolor}%
\usepackage{textcomp}%
\usepackage{manyfoot}%
\usepackage{booktabs}%
\usepackage{algorithm}%
\usepackage{algorithmicx}%
\usepackage{algpseudocode}%
\usepackage{listings}%

\usepackage[utf8]{inputenc} 
\usepackage[T1]{fontenc}    
\usepackage{hyperref}       
\usepackage{url}            
\usepackage{microtype}      
\usepackage{subcaption}
\usepackage{bm}
\usepackage{listings}       

\theoremstyle{thmstyleone}%
\theoremstyle{thmstyletwo}%

\theoremstyle{thmstylethree}%
\newtheorem{definition}{Definition}%

\begin{document}

\def\setmathchar#1{\ifmmode#1\else\(#1\)\fi}
\def\setformath#1{\ifmmode%
					  {\mathchoice{\mbox{#1}}%
								  {\mbox{#1}}%
								  {\mbox{\scriptsize#1}}%
								  {\mbox{\tiny#1}}}%
				  \else{#1}%
				  \fi}
\def\keyword#1{\setformath{\texttt{#1}}}

\newcommand{\TheToolName}{TOOLNAME}
\newcommand{\toolname}{\keyword{\TheToolName}}
\renewcommand{\algorithmicrequire}{\textbf{Input:}}
\renewcommand{\algorithmicensure}{\textbf{Output:}}
\algnewcommand\algorithmicforeach{\textbf{for each}}
\algdef{S}[FOR]{ForEach}[1]{\algorithmicforeach\ #1\ \algorithmicdo}
\algdef{S}[FOR]{ForEachParallel}[1]{\algorithmicforeach\ #1\ \algorithmicdo \textbf{ in parallel}}

\newcommand{\eat}[1]{}

\title[Article Title]{
    Rule-Based Explanations
    for Retrieval-Augmented LLM Systems
}


\author*[1]{\fnm{Joel} \sur{Rorseth}}\email{jerorset@uwaterloo.ca}
\author[2]{\fnm{Parke} \sur{Godfrey}}\email{godfrey@yorku.ca}
\author[1]{\fnm{Lukasz} \sur{Golab}}\email{lgolab@uwaterloo.ca}
\author[3]{\fnm{Divesh} \sur{Srivastava}}\email{divesh@research.att.com}
\author[2]{\fnm{Jarek} \sur{Szlichta}}\email{szlichta@yorku.ca}


\affil*[1]{
\orgname{University of Waterloo}, \orgaddress{
\city{Waterloo}, 
\state{Ontario}, \country{Canada}}}

\affil[2]{
\orgname{York University}, \orgaddress{
\city{Toronto}, 
\state{Ontario}, \country{Canada}}}

\affil[3]{
\orgname{AT\&T Chief Data Office}, \orgaddress{
\city{Bedminster}, 
\state{New Jersey}, \country{USA}}}


\abstract{
If-then rules are widely used to explain
machine learning models;
e.g., ``if employed = no, then loan application = rejected\@.''
We present the first proposal to apply rules to explain
the emerging class of large language models (LLMs) with
retrieval-augmented generation (RAG).
Since RAG enables LLM systems to incorporate retrieved
information sources at inference time, rules linking the
presence or absence of sources can explain output
provenance;
e.g., ``if a Times Higher Education ranking article is
retrieved, then the LLM ranks Oxford first\@.''
To generate such rules, a brute force approach would
probe the LLM with all source combinations and check
if the presence or absence of any sources leads to
the same output.
We propose optimizations to speed up rule generation,
inspired by Apriori-like pruning from frequent itemset
mining but redefined within the scope of our novel problem.
We conclude with qualitative and quantitative experiments
demonstrating our solutions' value and efficiency.
}




\maketitle


\section{Introduction}

Advances in large language models (LLMs) have yielded
increasingly sophisticated artificial intelligence (AI)
systems capable of accomplishing a wide variety of tasks.
These advances are enabled not only by the increasing size
and architectural complexity of the models, but also by
innovations in how LLMs incorporate knowledge.
One such prominent innovation is
\emph{retrieval-augmented generation} (RAG), by which the
static, pre-trained knowledge of an LLM is augmented with
dynamic, arbitrary knowledge retrieved at
inference time.
Although RAG enables more control over an LLM's data,
the provenance of information in the LLM's response is
further complicated as a result.

The emerging field of
\emph{explainable artificial intelligence} (XAI) has been
addressing issues related to understanding the behavior of
complex models, including LLMs.
Many XAI methods focus on the simpler problem of
\emph{local} explanation, which attempts to explain
specific model predictions rather than general model
behavior.
One popular local XAI methodology is
\emph{feature attribution}, which explains a model output
by scoring the relative importance of the input features.
For RAG systems, context attribution can assign scores to
each piece of information retrieved by an
LLM system at inference time \citep{contextcite}.
However, feature attribution is not actionable
(e.g., to achieve recourse).

\emph{Counterfactual} explanations address this problem,
explaining a model's output by citing a minimal input
perturbation that would induce an alternative output
\citep{wachter2017counterfactual}.
In RAG, a counterfactual explanation might reveal that the
LLM would have generated a different answer if some piece
of information were not provided or retrieved
at runtime \citep{RAGE}.
However, there may be many minimal counterfactuals for a
given prediction, which may overwhelm end users.
Explanations using if-then rules mitigate these problems
by summarizing patterns observed across a model's
predictions~\citep{anchors, MACHA2022101209, lore}, but
have not yet been considered in the context of
LLMs with RAG.

We fill this research gap and present a
rule framework for explaining the output of black-box
RAG systems using rules that formulate input predicates
based on the \emph{presence} of RAG sources provided
to the LLM.
Let us illustrate using a healthcare example.
Suppose that a clinic is using a RAG system connected to a
medical database, and a medical research assistant working
at the clinic asks the LLM, ``What treatment is most
effective for Long COVID fatigue\@?''
Here, a rule could reveal that ``Whenever `Document 123'
is included in the RAG sources,
the LLM always recommends an unsafe Long COVID treatment\@.''
(A rule's antecedent could likewise be
when a given document is \emph{omitted from} the RAG sources.)
This rule helps curators of the medical database flag
documents such as Document 123 that, unbeknownst to them,
advocate for unsafe treatments.
In this case, Document 123 may advocate for a treatment
that was thought to be effective until recently,
while newer studies have shown it is both ineffective and
unsafe for certain demographics.
Our framework empowers developers, auditors, policy makers,
and other end users to seek rules for \emph{any}
identifiable output condition, such as responses that are incorrect,
contain misinformation, or reflect negative sentiment \citep{han2024medical}.

Our main contributions are as follows.

\begin{enumerate}
\item
    We propose a \textbf{rule-based explanation}
    formulation for RAG, the first of its kind.
    We offer two variants that predicate on the basis of
    \emph{retaining} or \emph{omitting} features (RAG sources).
    A rule explains the output returned by an LLM by
    identifying conditions that consistently hold in
    practice whenever certain features are retained or omitted.
\item
    We design \textbf{efficient search algorithms} to mine
    rule-based explanations, exploiting intuitions of rule
    validity to prune the search space and enable early termination of the search.
\item
    We conduct an \textbf{experimental evaluation} of our
    explanation search algorithms, 
    presenting
    case studies showing the value of rule-based explanations
    of RAG systems for question answering,
    and highlighting the efficiency
    improvements due to our pruning techniques.
\end{enumerate}

The rest of the paper is structured as follows:
Section \ref{sec:relatedwork} reviews related work;
Section \ref{sec:formulation} introduces our rule formulation;
Section \ref{sec:algorithms} illustrates our rule mining algorithms;
Section \ref{sec:eval} presents an experimental evaluation;
and
Section \ref{sec:conclusion} concludes with future research directions.

\section{Related Work}
\label{sec:relatedwork}

In the XAI literature, many methods have been proposed
to explain individual model predictions,
collectively known as \emph{local} explanations.
Many local methods adopt an approach known as
feature attribution, exemplified by methods such as
\keyword{LIME}\ \citep{ribeiro2016should} and \keyword{SHAP}\ \citep{shap},
which explain a model by scoring input features
according to their relative influence on the model's
output.
Though intuitive, these explanations are not directly
actionable, and have been shown to be potentially
unreliable, misleading, and manipulable
\citep{Kindermans2019, advlime:aies20, shap-edu-problems}.
In contrast, counterfactual explanations explain a
model by perturbing the input features in a minimal
way to change the original output
\citep{wachter2017counterfactual, verma2024counterfactual};
e.g., a declined loan application would have been approved
if the applicant's income were 20 percent higher. 
Counterfactual perturbations aim to produce actionable
insights that facilitate recourse, but many valid
counterfactuals may exist, potentially contradicting
each other and overwhelming users.
Some methods therefore seek to sample diverse counterfactual
perturbations or prioritize those deemed more
feasible \citep{mothilal2020dice, NEURIPS2020_c2ba1bc5}.

At the cost of increased computational complexity,
rules mitigate the above problems by summarizing
patterns observed across a model's predictions
\citep{anchors, MACHA2022101209, lore, rudinrules};
e.g., ``if employed = no, then loan application = rejected,''
regardless of the values of the other features.
Recent works have established that rules essentially
subsume counterfactuals, forming a duality,
since both make assertions over a model's input
features \citep{geng2022computing, guidotti2024stable}.
Rule-based explanations have been defined as
conjunctions of \emph{predicates} over
the features of an input, for which any matching input
always leads to the model producing the same output
\citep{rudinrules}.
Several methods, such as \keyword{Anchors}\ \citep{anchors}, have
applied similar definitions for the purposes of XAI.
In contrast to our work, their rules are approximate
(i.e., hold with high confidence), not exact,
and do not provide an interface to instantiate their
rules' input predicates for modern LLMs or RAG systems.

These rule formulations build implicitly upon
\emph{association rule} formulations established
in the rule mining literature
\citep{rules-1993, apriori, most-interesting-rules}.
These formulations gave rise to many influential
\emph{rule mining} algorithms that search
for frequent itemsets and valid association rules,
most notably the Apriori algorithm \citep{apriori} and
more-efficient successors like Eclat \citep{eclat}
and FP-growth \citep{fpgrowth}.
The rule mining methods we present in this work
take algorithmic inspiration from Apriori, but have
several key modifications for a fundamentally
different problem: explaining AI model predictions.
While Apriori iterates over observed \emph{itemsets},
we instead iterate over the powerset of possible model
inputs (features), eliminating any concept of
\emph{support} (see Definition \ref{def:rule_support})
in rule validity.
This enables us to seek sets that satisfy arbitrary
predicates over model outputs, as opposed to those
observed with some minimum \emph{frequency}, as in the original Apriori algorithm.
Our algorithms also leverage the concept of
\emph{monotonicity} (i.e., the Apriori property)
to prune the search space, but we propose a new
definition that operates on model output predicates
rather than itemset support.

Naturally, many recent XAI works have focused on
explaining LLMs, RAG, and other related phenomena,
adapting general XAI methodologies as needed.
\emph{Global explanation} disciplines such as
mechanistic interpretability attempt to identify
low-level circuits that form among LLM neurons
\citep{autocircuits, NEURIPS2024_a8f7d43a, NEURIPS2024_ad217e0c}.
However, these insights are intended primarily
for scientific (i.e., expert) understanding.
One intuitive explanation format for LLMs and RAG
systems are \emph{citations}, which can be
generated at inference time using
specialized training and fine-tuning methods
\citep{sourceaware2024, aly-etal-2024-learning}
or post-hoc using various citation recommendation
strategies
\citep{farber2020citation, maheshwari-etal-2025-citefix, ceg}.
However, these methods are subject to the critical
limitation of hallucination
\citep{agrawal-etal-2024-language}, as their
predictive approach to generating citations can be
highly inaccurate, even with state-of-the-art models
\citep{citeme, cao-wang-2024-verifiable, byun-etal-2024-reference, day2023preliminary, alce}.
Meanwhile, recent works have adapted feature attribution
\citep{contextcite} and counterfactual methods
\citep{RAGE} to explain source importance in RAG,  assigning importance weights to the retrieved sources or  generating counterfactuals by removing some sources to change the LLM's output, respectively.
In contrast to our rule-based approach, these recent
methods are prone to the same limitations as their
general methodologies, and may fail to produce concise and actionable
insights that are robust and guaranteed.

\section {Rule-Based Explanation Formulation}
\label{sec:formulation}

In this section, we introduce our novel rule-based
explanation formulation.
We start by reviewing the design space.
We then provide a mathematical formalization of our
rule-based explanations, clarifying how the hierarchical
nature of our rules' validity leads to complementary
variants that either \emph{retain} or \emph{omit} inputs.
Finally, we explain how rule validity can be defined
recursively, and interpreted using a lattice
data structure.

\subsection{Rule Design}
\label{subsec:rule-design}

In the data mining literature,
a \textit{rule} expresses a conditional
``if-then'' relationship where satisfying the ``if'' part
(the \textit{antecedent}) implies that the ``then'' part
(the \textit{consequent}) is also satisfied
\citep{most-interesting-rules}.
One intuitive adaptation for rules describing a
prediction $y = M(x)$ might define the
antecedent in terms of the input $x$ to the model $M$,
and the consequent in terms of the resulting output $y$.
This would yield rules of the form
``if $x$ then $y$''.
We adopt this general orientation due to its
clear alignment with the primary goal of local XAI:
understanding the effect of model inputs on
model outputs.
This orientation has also been adopted in recent XAI works,
wherein rules explain model predictions by uncovering
significant patterns in said predictions.
In this XAI setting, rules are better known as
\emph{rule-based explanations}.

\begin{definition}[Rule-Based Explanation]
\label{def:rule_based_expl}
A rule-based explanation is a
conjunction of predicates over the features of some model
$M$, where for any input $x^*$ satisfying
\textit{all} predicates, the model \textit{always}
generates some target output $y^*$
\citep{geng2022computing, rudinrules}.
\end{definition}

In the XAI setting, the antecedent is an assertion over
a conjunction of input predicates, and the consequent is
an assertion over a model output.
In the literature, rules are often expressed and mined
in terms of some set of \emph{observed} instances,
which would correspond to a model's \emph{predictions}
in our setting.
Given a set of observed instances, a rule may
be judged by the number of observed instances that
match both sides of the rule, or by the proportion
of instances whose consequent is satisfied when the
antecedent is satisfied.
These empirical measurements, which are often used as
a proxy for rule quality, are known formally as the
\emph{support} and \emph{confidence} of a rule, respectively.

\begin{definition}[Rule Support]
\label{def:rule_support}
Given a set of observed instances,
the support of a rule is the proportion of instances
that satisfy the rule.
\end{definition}

\begin{definition}[Rule Confidence]
\label{def:rule_confidence}
Given a set of observed instances,
the confidence of a rule is the conditional
probability that its consequent is satisfied,
measured over all instances in the set
that satisfy its antecedent.
\end{definition}

In settings that observe such instances, rules are
considered more informative when they have higher
support (i.e., the rule is more likely to implicate
an instance) and higher confidence (i.e., the rule
is more likely to hold).
In this work, we do not seek rules that characterize
observed data, and thus do not presume a setting
wherein a set of observed instances is available.
Instead, we seek rules that characterize the full
model input space, and therefore operate over the
set of all possible antecedents.
It follows that the classical notion of support does
not apply in our setting, though it is nonetheless equal
to the reciprocal of the number of possible
antecedents.
Furthermore, we require that all valid rules hold with
certainty (i.e., confidence of 100\%), but note that
future work could build upon existing rule methods
that guarantee minimum confidence levels
\citep{anchors}.

In practice, ambiguity still remains when applying this
formulation, as different models, features, problems, and
user preferences necessitate different predicates.
We now propose a concrete implementation for the
RAG setting.

\textbf{Antecedent:}
For our rules, the choice of antecedent input
predicate(s) is critical, as it will determine their
explanatory effectiveness and the computational
complexity of mining them.
Ideally, for maximum explanatory power, the antecedent
of a rule should implicate as many inputs as possible
(i.e., wide coverage).
For example, the antecedent ``If the LLM is given
a \emph{set} of sources that \emph{includes}
`Document 123'\,'' covers more of the feature space than
``If the LLM is given `Document 123'\,''.
Likewise, the chosen antecedent predicate must also be
compatible with and reasonable for the modality of the
input features.
While categorical features may be constrained to 
specific values (e.g., marital status is ``single''),
and numeric features to a range of values
(e.g., credit score is $\geq 600$),
the ideal definition for the \textit{textual} inputs of
language models is less clear.

Drawing inspiration from XAI literature, we contend that
\emph{feature ablation} (i.e., removing text segments) is a
suitable perturbation strategy for explaining language models,
and thus formalize antecedent predicates based on feature
\textit{presence}.
In the XAI literature, ablation is a well-founded
strategy for inferring feature importance
\citep{hameed:basedxai:2022}, and has seen
applications in works that explain language models
\citep{RAGE,contextcite} and
models generally \citep{shap}.
We assume that the user partitions the input $x$
into two components $x = (\mathbf{s}, \mathbf{c})$,
where $\mathbf{s}$ represents some
\textit{feature subset} of interest
$\mathbf{\mathbf{s}} = \{ s_1, s_2, \dots, s_n \}$
derived from the input (e.g., RAG document sources).
In our rule formulation, the antecedent will be
defined in terms of the presence (or similarly,
the absence) of inputs in $\mathbf{s}$, while the
remaining inputs in $\mathbf{c}$ are ignored.
By separating $\mathbf{s}$ and
$\mathbf{c}$ (i.e., $y = M(\mathbf{s'}, \mathbf{c})$),
we enable the user to focus their rules
and explanations on small, interesting subsets of the
input feature space, while reducing the complexity
of our rule mining algorithms.

\textbf{Consequent:}
We contend that the ideal choice of consequent
is ultimately dictated by the user and their motivation
for seeking rules.
Therefore, the consequent of our rules asserts only that
the output meets some \emph{target condition} defined
explicitly by the user.
In essence, the target condition is an arbitrary output
predicate that could test for any identifiable property
(e.g., presence of misinformation,
validity of in-text citations).
We expect the user to define an
\emph{output predicate function}
$O: \mathcal{Y} \rightarrow \{0,1\}$
that returns 1 when a model output $y \in \mathcal{Y}$
meets these conditions of interest.
By parameterizing the consequent, we empower the
user to seek rules that implicate not just the presence
of a specific output (as is common in many XAI and
rule-based works), but more sophisticated
output characteristics
(e.g., presence of hallucination, bias, sentiment).

To define an output predicate function $O$, the user
must effectively implement a classifier that inspects
an arbitrary LLM output and judges whether it meets
the conditions of interest.
A simple predicate might compare the answer for
equality against some \emph{target} answer, returning
1 if the LLM answer is equal to the target answer
(and 0 otherwise).
A more specific predicate might analyze the LLM answer
for \emph{correctness}; however, this would only be
possible in settings where the correct answer is known
upfront (e.g., benchmarking).
For nuanced semantic assertions (e.g., judging citation
validity), or those that must be robust to verbose or
unstructured LLM outputs (e.g., spotting inappropriate
content), a predicate function could adopt an
``LLM-as-a-judge'' implementation.
Given detailed instructions and necessary context
(e.g., ground truth information), an LLM can be a highly
effective proxy for human judgment of LLM responses
\citep{chiang-lee-2023-large, judging-llm-judge};
however, stochasticity introduced during the decoding
process (e.g., token sampling) can compromise
reproducibility.

\begin{figure}[t]
\centering
    \includegraphics[width=\textwidth]{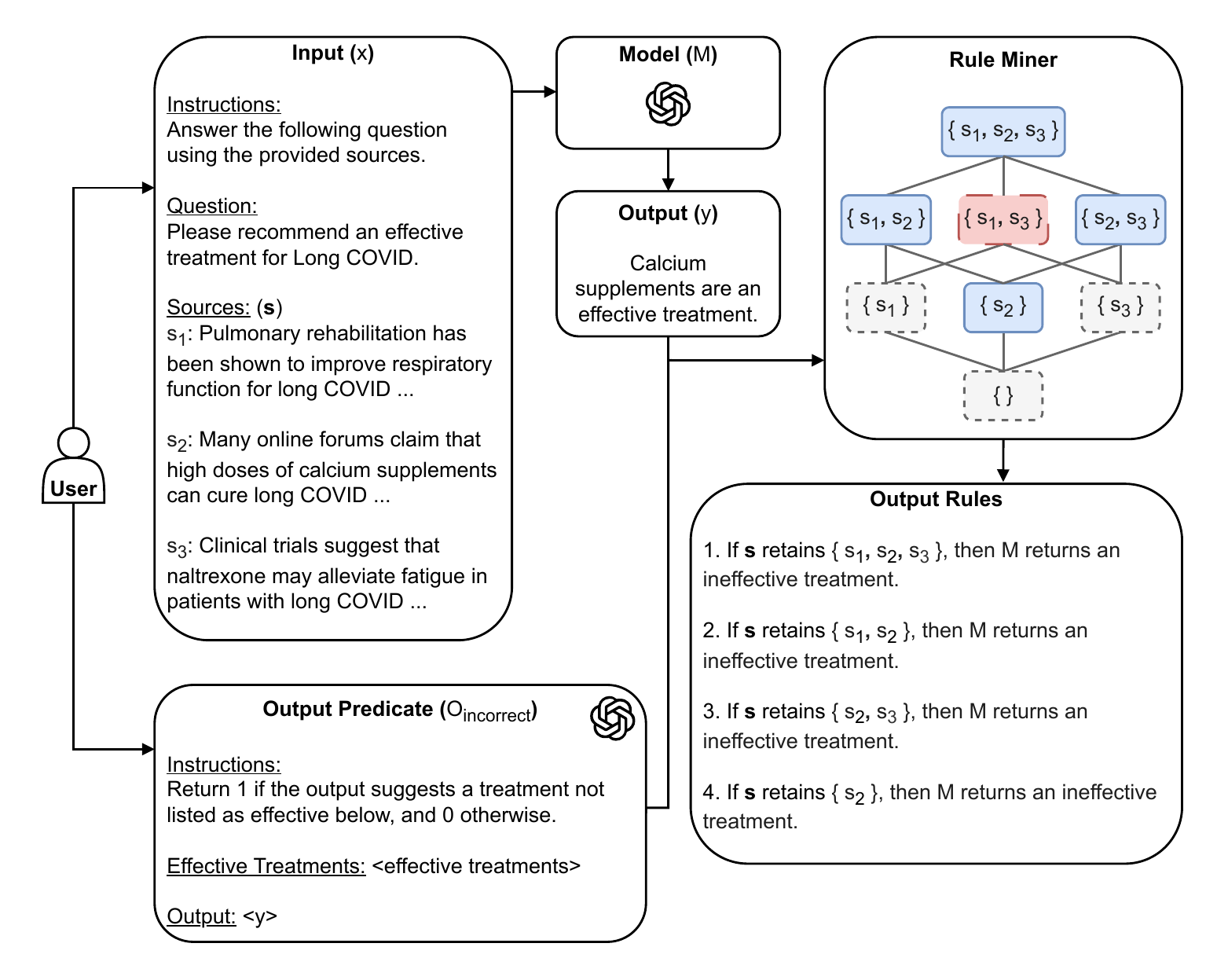}
    \caption{
    This diagram illustrates an example where the user
    asks a RAG system to recommend an effective treatment
    for Long COVID, given three retrieved sources.
    The rules generated by our algorithm (formalized in
    Section \ref{sec:algorithms}) indicate that retaining
    source $s_2$ consistently leads the LLM to recommend
    treatments known to be ineffective.
    }
    \label{fig:flow}
\end{figure}

\textbf{Example:}
Consider the following running example, illustrated in
Figure \ref{fig:flow}.
Suppose a user wants to debug incorrect predictions
made by an LLM that is being used to suggest
treatments for medical illnesses.
The LLM is configured to use RAG,
meaning that the LLM's pre-trained knowledge is
supplemented with external knowledge sources
$\mathbf{s}$ (relevant medical documents) to help
ground its answers.
The user asks the LLM to recommend an effective
treatment for Long COVID, and is investigating why
an incorrect answer ``calcium supplements'' is
consistently returned.
Let $x$ represent the corresponding prompt
submitted to the LLM $M$ by the user, and $y$
represent its incorrect response, as illustrated in
Figure \ref{fig:flow}.
All inputs in the prompt other than $\mathbf{s}$
(i.e., the question and LLM instructions) are captured
in $\mathbf{c}$, and will not be part of any rules.

Suppose that the user wants to know whether
any patterns emerge among the top-3 retrieved medical
documents (sources) when ineffective
treatments (i.e. incorrect answers) are observed.
To make this intent explicit, the user
defines a function $O_{incorrect}$ that returns 1 if
$y$ recommends a treatment not known to be effective
for the indicated condition, and 0 otherwise.
The user implements $O_{incorrect}$ by constructing a
prompt to be posed to a separate LLM.
It contains a list of treatments known to be clinically
effective for Long COVID, and asks it to return 1 if
the observed response suggests a treatment not in
this list.
This definition is illustrated in Figure \ref{fig:flow},
which shows how the input $x$ and output predicate
$O_{incorrect}$ (along with $M$ and $y$) are
subsequently provided to our rule mining algorithm.
The algorithm proceeds by considering all possible
combinations of the three medical documents, performing
inference for certain combinations (as necessary),
then finally deriving patterns across combinations
that satisfied $O_{incorrect}$.

As shown in Figure \ref{fig:flow}, one rule found to
be valid under $O_{incorrect}$ asserts that
``if $\mathbf{s}$ retains all sources in $\{s_2\}$,
then $M$ returns an ineffective treatment\@.''
Other valid rules are found, but they are simply
more-specific versions of this rule
(i.e., implicating supersets of $\{s_2\}$).
If the user were instead interested in patterns that
lead to a correct answer, they could define a
reciprocal predicate $O_{correct}$, potentially
producing a symmetric rule such as
``if $\mathbf{s}$ omits all sources in $\{s_2\}$,
then $M$ returns an effective treatment\@.''
In either case, the rule suggests that the presence of
$s_2$ is important for the LLM to recommend an
ineffective treatment.

\subsection{Rule Formulation}
\label{subsec:rule-formulation}

The concepts of presence and absence
have a reciprocal relationship.
Suppose that we have a subset of the input set,
$\mathbf{s'} \subseteq \mathbf{s}$, and that we aim to
characterize the contents of $\mathbf{s'}$ w.r.t.
$\mathbf{s}$.
\emph{Retaining} all elements from $\mathbf{s}$ that
appear in $\mathbf{s'}$ is equivalent to
\emph{omitting} all elements from $\mathbf{s}$ that
appear in the complement of $\mathbf{s'}$,
$(\mathbf{s} \setminus \mathbf{s'})$.
Conversely, omitting all elements from $\mathbf{s}$
that appear in $\mathbf{s'}$ is equivalent to retaining
all elements in $(\mathbf{s} \setminus \mathbf{s'})$.
This duality admits two variants of our rule-based
explanations: \emph{retention} and \emph{omission}
rules.

\begin{definition}[Retention Rule]
\label{def:retention_rule}
A retention rule for some output predicate $O$ 
and implicated input subset
$\mathbf{s'} \subseteq \mathbf{s}$
guarantees that
$\forall \mathbf{t} \in \mathcal{P}(\mathbf{s}),\ \mathbf{s'} \subseteq \mathbf{t} \Rightarrow O(M(\mathbf{t}, \mathbf{c})) = 1$.
\end{definition}

\begin{definition}[Omission Rule]
\label{def:omission_rule}
An omission rule for some output predicate $O$ 
and implicated input subset
$\mathbf{s'} \subseteq \mathbf{s}$
guarantees that
$\forall \mathbf{t} \in \mathcal{P}(\mathbf{s}),\ (\mathbf{t} \cap \mathbf{s'} = \emptyset) \Rightarrow O(M(\mathbf{t}, \mathbf{c})) = 1$.
\end{definition}

A retention rule identifies a subset
of the input set that, when retained, always satisfies
the output predicate.
Similarly, an omission rules identifies a subset that,
when omitted, always satisfies the output predicate.
Retention and omission rules will often correlate with
outputs that satisfy reciprocal (i.e., complementary)
output predicates.
Therefore, users will often use two separate,
reciprocal output predicates when explaining a single
input (e.g., $O_{correct}$ and $O_{incorrect}$).
Generally, both rule types identify critical inputs
in $\mathbf{s}$ whose presence is necessary to ensure
some condition holds.

Revisiting our RAG example from
Figure \ref{fig:flow}, recall that the user can obtain
rules of two symmetric types, either in terms of answer
correctness or incorrectness.
In this example, the retention rule formulation aligns
with $O_{incorrect}$, and the omission rule formulation
aligns with $O_{correct}$.
That is, the aforementioned retention rule
under $O_{incorrect}$ asserts that, across all
combinations of sources in $\mathbf{s}$, the LLM's answer
is always incorrect when the combination retains all
sources in $\mathbf{s'} = \{s_2\}$.
Likewise, the complementary omission rule under
$O_{correct}$ would assert that the answer is correct
anytime the combination omits $\mathbf{s'} = \{s_2\}$.

To guarantee these assertions for a subset
$\mathbf{s'} \subseteq \mathbf{s}$, one must verify that
the output predicate remains satisfied for all possible
subsets of $\mathbf{s}$ wherein the inputs in
$\mathbf{s'}$ are retained (or omitted, respectively).
Following the sources $\mathbf{s}$ defined in
Figure \ref{fig:flow}, where $|\mathbf{s}| = 3$,
validating either $\{ s_2 \}$ rule would require evaluating
the LLM output corresponding to each subset in
$T = \{ \{s_2\}, \{ s_1, s_2 \}, \{ s_2, s_3 \}, \{ s_1, s_2, s_3 \} \}$.
To validate the $\{ s_2 \}$ retention rule
against $O_{incorrect}$, one would need to ensure that
$\forall \mathbf{t} \in T$,
$O_{incorrect}(M(\mathbf{t}, \mathbf{c})) = 1$.
The ``Rule Miner'' module in Figure \ref{fig:flow}
depicts these as blue nodes with a solid border, which
indicates their validity.
To validate the $\{ s_2 \}$ omission rule against
the applicable predicate $O_{correct}$, one must
ensure that
$\forall \mathbf{t} \in T$,
$O_{correct}(M(\mathbf{s} \setminus \mathbf{t}, \mathbf{c})) = 1$.
In the next subsection, we formalize a definition of
rule validity using recursion, which informs the
design of our novel rule search algorithms.

\subsection{Rule Validity}

Under our definitions, the validity of a
rule can be defined recursively.
A retention rule implicating $\mathbf{s'}$ is valid
iff the predicate $O$ is satisfied for all supersets
of $\mathbf{s'}$.
Likewise, for an omission rule $\mathbf{s'}$,
$O$ must be satisfied for all subsets of its
complement $(\mathbf{s} \setminus \mathbf{s'})$.
Without loss of generality, the validity of a rule
requires output predicate satisfaction for all
\emph{ancestors} of $\mathbf{s'}$, including
$\mathbf{s'}$ itself.
The set of ancestors can be defined recursively by
repeatedly applying the \emph{parent} relation,
starting from $\mathbf{s'}$ and continuing until no
further parents exist.
Thus, valid rules have an intrinsically recursive
relationship, as their validity depends on the
total validity of all ancestors.

\begin{definition}[Rule Validity]
\label{def:rule_validity}
Let $P_{\mathbf{s'}}$ denote the set of
parents of $\mathbf{s'}$, i.e., the immediate
supersets of $\mathbf{s'}$ (for a retention rule) or
the immediate subsets of
$(\mathbf{s} \setminus \mathbf{s'})$ (for an
omission rule).
We can define rule validity as a recursive predicate
$V(\mathbf{s'}) = O(\mathbf{s'}) \wedge \forall \mathbf{t} \in P_{\mathbf{s'}}, V(\mathbf{t})$.
\end{definition}

By definition, all ancestors must be validated
\emph{before} validating a descendant.
A rule mining algorithm must first validate those
rules that retain or omit the \emph{most} inputs
(i.e., fully recurse) before unwinding the call stack
to consider rules that retain or omit fewer inputs.
Since a rule implicates all ancestor rules whose
representative subset is a superset of its own,
rules representing smaller subsets will implicate more
rules than those representing larger, meaning that
descendant rules have greater coverage than ancestor rules.
Under this interpretation, descendant rules can be seen
as \emph{subsuming} ancestor rules as the latter
have less coverage (implicate fewer feature
combinations).

\begin{definition}[Rule Subsumption]
\label{def:rule_subsumption}
A rule $r_1$ subsumes another rule $r_2$ iff
every predicate in $r_1$ also appears in $r_2$.
\end{definition}

Moreover, rules that retain or omit \emph{fewer} inputs
(i.e., descendants) are considered to have greater
explanatory power,
since by definition, their output predicate holds over
more inputs (i.e., greater coverage).
Rules
that implicate smaller feature subsets $\mathbf{s'}$
are considered to be more \emph{minimal}.

\begin{definition}[Rule Minimality]
\label{def:rule_minimality}
A rule $r_1$ is minimal iff there exists no rule $r_2$
such that $r_2$ subsumes $r_1$ and $r_2 \neq r_1$.
\end{definition}

In this work, we seek minimal rules, but operate under
the understanding that less-minimal rules must be
validated first.
We acknowledge an inherent trade-off between
rule minimality and algorithmic efficiency,
as finding \emph{more} minimal (higher coverage)
rules ultimately requires validating \emph{more} rules.
Our algorithms (introduced in
Section \ref{sec:algorithms}) ultimately prioritize
minimality, exhausting all rule candidates and
guaranteeing 100\% confidence.
However, to eliminate redundancy, non-minimal rules
can be discarded before being returned to the user.

\begin{figure*}[htbp]
\centerline{
    \includegraphics[width=1.0\textwidth]{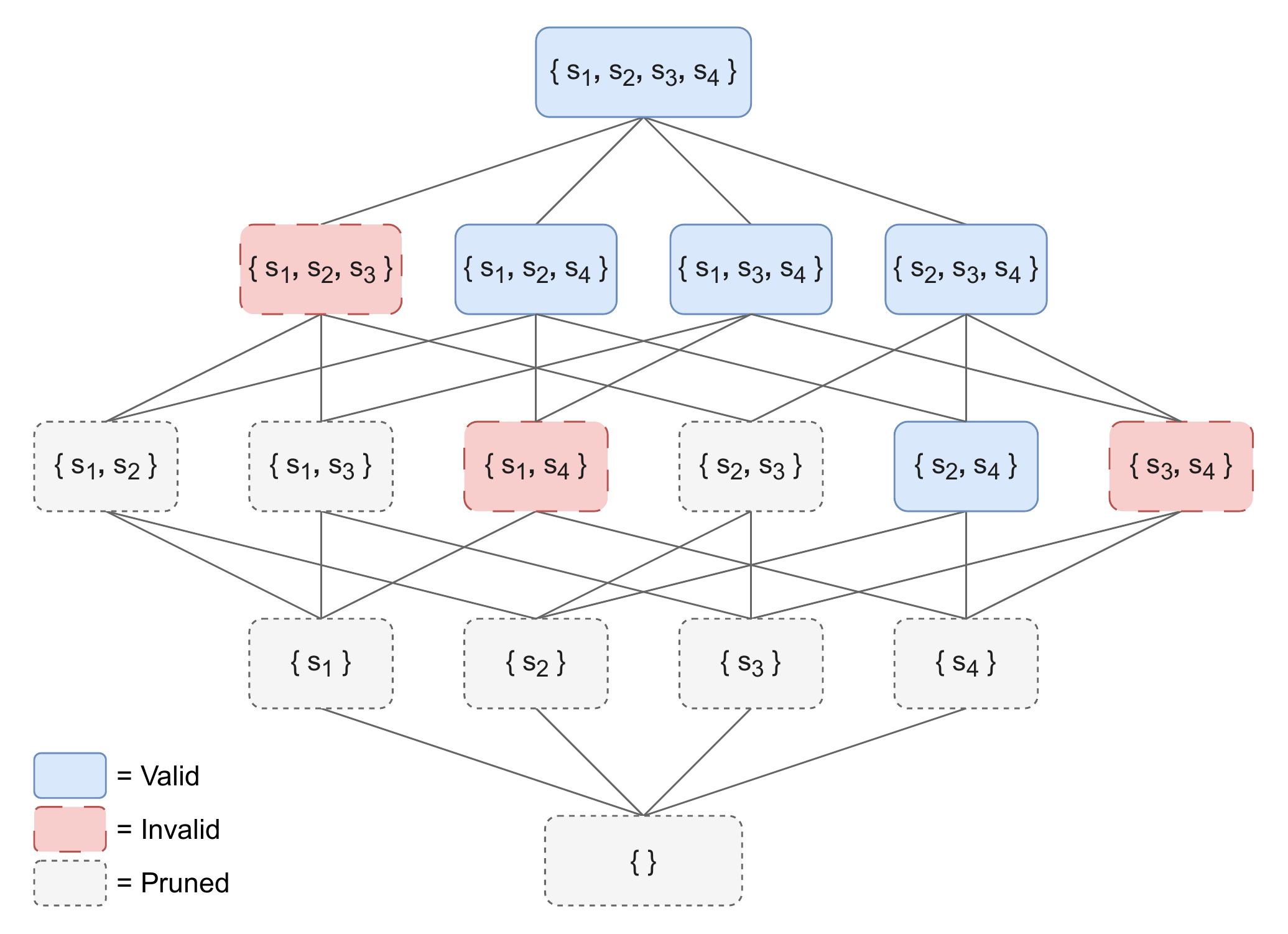}
}
\caption{
A lattice representing an input set
$\mathbf{s}$ where $|\mathbf{s}| = 4$.
The node coloring illustrates an example run for
the top-down breadth-first search performed by
Mono Rule Miner
(introduced in Section \ref{subsec:mono}),
which discovers valid rules (blue with solid border),
invalid rules (red with long dashes), and candidates
that can be pruned (grey with short dashes).
}
\label{fig:lattice}
\end{figure*}

The full hierarchy of a given rule is a subset of
the power set $\mathcal{P}(\mathbf{s})$.
The power set, which contains \emph{all} candidate
rules that implicate $\mathbf{s}$, forms a
data structure known as a \emph{lattice}
\citep{abiteboul1995foundations},
illustrated in Figure \ref{fig:lattice}.
Each of the $2^{|\mathbf{s}|}$ subsets in
$\mathcal{P}(\mathbf{s})$ is a rule candidate; each
node in the lattice is a candidate, and each edge
delineates a subset / superset relationship between
two candidates.
The $\mathcal{P}(\mathbf{s})$ lattice constitutes the
full search space for any rule mining algorithm under
our formulation, and it contains $2^{|\mathbf{s}|}$
candidate rules.
In the following subsections, we present two rule
mining algorithms that efficiently navigate this
lattice in search of \emph{all} valid rules.

\section {Rule Mining Algorithms}
\label{sec:algorithms}

In this section, we introduce two search algorithms%
\footnote{%
    Algorithm implementations are available at
    \url{https://github.com/joelrorseth/rag-llm-rule-miner}.
\label{footnote:algorithms}
}
capable of mining our rule-based explanations,
each with different advantages.
First, we introduce the Mono Rule Miner,
which can mine either retention or omission rules in a
single pass.
Second, we introduce the Dual Rule Miner,
which can mine both rule types simultaneously in a single
pass.
Mono is more efficient in settings where a single rule
type is sufficient, and can exploit greater
search space pruning.
For settings where patterns of both types
are sought and compared, Dual is often more efficient
than executing two Mono passes.

\subsection{Mono Rule Miner}
\label{subsec:mono}

As discussed in Section \ref{subsec:rule-formulation},
users will request rule-based explanations after
obtaining a model output $y = M(\mathbf{s}, \mathbf{c})$.
To define the desired rule, the user will specify whether
they seek retention or omission rules, and will supply a
definition for an output predicate $O$.
The goal of our algorithms is to discover all valid rules
that implicate subsets of $\mathbf{s}$.
This necessitates a thorough (and potentially exhaustive)
search through the lattice, and subsumes the simpler
problem of validating a single candidate rule.

As exemplified in Figure \ref{fig:lattice}, the lattice
contains $|\mathbf{s}| + 1$ \textit{levels} that group
rule candidates by their size (i.e., $|\mathbf{s'}|$ for
a rule based on $\mathbf{s'} \subseteq \mathbf{s}$).
Since rule validity is defined recursively w.r.t. all
ancestors, we must generally navigate the lattice
top-down to validate rules.
Due to the recursive definition, validating any single
rule can be accomplished using a
\textit{depth-first search} (DFS), starting from the
rule in question and recursing to all its ancestors in
the lattice.
Since we seek \emph{all} valid rules, simply executing
DFS for each candidate rule (top-down)
would incur significant repeated computation.
Recomputation can be avoided by caching rule validity
during a top-down traversal, reducing runtime
complexity to $\mathcal{O}(2^{|\mathbf{s}|})$,
but caching would impose an expensive
$\mathcal{O}(2^{|\mathbf{s}|})$ space complexity.

Instead, we propose a top-down
\textit{breadth-first search} (BFS) implementation
that finds \emph{all} valid rules of a specific type
in a \emph{single} pass.
We call this our \emph{Mono Rule Miner}.
The algorithm is detailed in
Algorithm \ref{algo:mono},
and its design is informed by two key observations.
The first is that our recursive definition
of rule validity admits a memory-efficient dynamic
programming solution.
The second observation is that the recursive definition
presents an opportunity for early termination when a
rule is found to be invalid.

\textbf{Dynamic Programming (Observation \#1):}
To justify our dynamic programming solution, we
observe that our recursive problem formulation is
already a dynamic programming problem, since the
validity of a rule is dependent on overlapping
subproblems originating in its ancestor rules.
Capitalizing on this, we can bound the number of
cached subproblems (i.e., rule validity records) to a
size that modestly improves over the naive
$\mathcal{O}(2^{|\mathbf{s}|})$.
Though the validity of a rule hinges
also on the validity of all its ancestors, it is
sufficient to hinge only on the validity of its
immediate \emph{parents}.
This is true because the remaining ancestors must
have already been validated when validating these
parents.
Therefore, we need only cache the validity of
two levels worth of rules at any time (the current
and previous level), resulting in a lower space
complexity of $\mathcal{O}{|\mathbf{s}| \choose l^*}$
where $l^* = \lfloor |\mathbf{s}| / 2 \rfloor$
(the index of the largest level).

\textbf{Early Termination (Observation \#2):}
Our second observation arises from our
definition of rule validity.
Since a rule's validity hinges on the validity of all
ancestors, then if a rule candidate is found to be
invalid, then \emph{none} of its descendants can
possibly be valid.
This property is analogous to the
\textit{downward closure property} (Apriori property)
from the data mining literature \citep{apriori},
and is key to ``pruning'' large subtrees of the
lattice search space.
Upon encountering an invalid rule, we can cache a
record of its invalidity and avoid evaluating (i.e.,
performing model inference for) the child rules
if and when we reach them during traversal.
Furthermore, once a level has no valid rules, the
entire search can terminate as all valid rules
must have been found.

\begin{algorithm}[t]
\caption{Mono Rule Miner}
\label{algo:mono}
\begin{algorithmic}[1]

\Require The input set $\mathbf{s}$
\Require The remaining inputs $\mathbf{c}$
\Require The output predicate $O$
\Require The model $M$
\Ensure The rule-based explanations
\vspace{10pt}

\State $R \gets \{\}$ \Comment{Initialize valid rule set}
\State $Z \gets \{\}$ \Comment{Initialize invalid node set}

\For{$l' \in [0, |\mathbf{s}|]$}
\label{linenum:mono:level_loop}

    \State $l \gets |\mathbf{s}| - l'$
    \Comment Compute current level size
    \label{linenum:mono:level_size}

    \State $S \gets \{ \mathbf{s'} \subseteq \mathbf{s} \mid |\mathbf{s}'| = l \}$
    \Comment Get all current level nodes
    \label{linenum:mono:level_nodes}

    \State $Z \gets \{ \mathbf{z} \subset \mathbf{z'} \mid \mathbf{z'} \in Z \land |\mathbf{z}| = l \}$
    \Comment{Get children of invalidated parents}
    \label{linenum:mono:level_invalid_parents}

    \State $Q \gets S \setminus Z$
    \Comment Prune invalid nodes in current level
    \label{linenum:mono:prune}

    \For{$\mathbf{s'} \in Q$}
    \label{linenum:mono:node_loop}
        
        \State $y \gets M(\mathbf{s'}, \mathbf{c})$
        \Comment Invoke the model
        \label{linenum:mono:inference}
    
        \If {$O(y) = 1$}
        \label{linenum:mono:predicate_check}
            \State $R \gets R \cup \{ \mathbf{s'} \}$
            \Comment{Record valid rule}
            \label{linenum:mono:record_valid}

        \Else
            \State $Z \gets Z \cup \{ \mathbf{s'} \}$
            \Comment{Record invalid node}
            \label{linenum:mono:record_invalid}
        \EndIf

    \EndFor
    
\EndFor

\State \Return $R$

\end{algorithmic}
\end{algorithm}

\textbf{Example:}
To illustrate how the the Mono Rule Miner works,
we will now explain how Algorithm \ref{algo:mono}
navigates the lattice in Figure \ref{fig:lattice}
to find all valid rules of a specific type.
Once provided with inputs
$\mathbf{s}, \mathbf{c}, O$ and $M$,
the algorithm will initialize caches for valid and
invalid rule candidates (nodes corresponding to subsets
of $\mathbf{s}$), then proceed
with a top-down level-order traversal
(line \ref{linenum:mono:level_loop}).
Supposing that $|\mathbf{s}| = 4$ as illustrated in
Figure \ref{fig:lattice}, the traversal begins by
gathering (line \ref{linenum:mono:level_nodes})
and looping through (line \ref{linenum:mono:node_loop})
subsets of size 4
(i.e., any $\mathbf{s'} \subseteq \mathbf{s}$ where
$|\mathbf{s'}| = 4$, meaning only $\mathbf{s}$).
Inside this loop, inference is performed on each of
these subsets (line \ref{linenum:mono:inference}),
and the predicate function $O$ is invoked with the
output to determine rule validity
(line \ref{linenum:mono:predicate_check}).
For this first subset $\{ s_1, s_2, s_3, s_4 \}$,
the output is found to satisfy $O$ (as indicated
in Figure \ref{fig:lattice})
and is added to the valid candidates cache.

The algorithm then proceeds to the next level
containing subsets of size three.
Since $\mathbf{s}$ is a valid rule and is the sole
parent of all subsets on this level, then each
subset \emph{may} be a rule,
and will be if it satisfies $O$.
As reflected in Figure \ref{fig:lattice},
all subsets are found to be valid rules except for
$\{ s_1, s_2, s_3 \}$, whose descendants can now
logically be ruled as invalid.
When the algorithm discovers that this subset is invalid,
only the subset itself is added to the
invalid candidates cache
(line \ref{linenum:mono:record_invalid}), as
descendants will be pruned dynamically upon
navigating to subsequent levels.
Proceeding to the next level with subsets of size 2,
the invalid candidates cache (containing only
$\{ s_1, s_2, s_3 \}$) is emptied and repopulated
(lines \ref{linenum:mono:level_invalid_parents}
and \ref{linenum:mono:prune})
with the \emph{children} of all members
($\{ s_1, s_2 \}$, $\{ s_1, s_3 \}$, $\{ s_2, s_3 \}$),
effectively pruning nodes from the subsequent
evaluation loop.

\textbf{Analysis:}
In the worst case, when \emph{all} $2^{|\mathbf{s}|}$
rule candidates are valid, the algorithm must make
$2^{|\mathbf{s}|}$ inference calls and use
$\mathcal{O}({|\mathbf{s}| \choose l^*})$ space.
However, in the opposite case, when no valid rules exist,
our real-time pruning short-circuits the algorithm,
requiring only a single inference call.
Otherwise, the complexity varies depending on the number
of valid rules, their location in the lattice, and many
other indirect factors.
These factors are analyzed empirically in
Section \ref{sec:eval}.

\subsection{Dual Rule Miner}
\label{subsec:dual}

As the name implies, the limitation of the
Mono Rule Miner is that only rules of a single type
are mined in its sole pass.
In settings where comprehensive tests
are demanded (e.g., LLM red teaming), obtaining rules
for a pair of reciprocal rule types enables deeper
understanding of the RAG sources and their influence.
With rules of two reciprocal types, a user can compare
whether a set of sources $\mathbf{s'}$ was sufficient,
both when retained \emph{and} omitted, to induce
an LLM output that meets a single target condition.
If $\mathbf{s'}$ is implicated in both types of rules,
then users can be more confident in its absolute
importance.
However if $\mathbf{s'}$ is implicated in only one
rule type, users can infer that its influence is
relative to the presence (or absence) of the other
sources.

To find rules of both types using Mono, each would
require a separate pass, and may utilize
different output predicates.
Despite potentially different predicates,
the underlying BFS still navigates the same lattice,
in the same order, for the same input set
$\mathbf{s}$, regardless of the chosen type.
Algorithmically, the key difference between the two
types stems from the \emph{interpretation} of the
``current subset;'' i.e., how it will be applied.
For the current subset $\mathbf{s'} \subseteq \mathbf{s}$,
a retention rule must \emph{retain} $\mathbf{s'}$,
but an omission rule must \emph{omit} $\mathbf{s'}$.

Guided by two additional observations (to follow),
we devise a strategy to navigate the lattice in search
of both types of rules, simply by
interpreting each subset in opposite ways.
We refer to the resulting algorithm as the
\emph{Dual Rule Miner}.
It proceeds with the same top-down BFS
lattice navigation as the Mono Rule Miner, finding
rules of both types; however, some efficiency is
sacrificed.
Specifically, the benefits of our early termination
(to reduce runtime complexity) and dynamic programming
caching (to reduce space complexity) introduced in
Section \ref{subsec:mono} are diminished.

\textbf{Mask Representation (Observation \#1):}
Recall from Section \ref{subsec:rule-formulation} that
presence and absence have a reciprocal relationship.
Omitting $\mathbf{s'}$ is equivalent to retaining its
complement w.r.t. $\mathcal{P}(\mathbf{s})$,
($\mathbf{s} \setminus \mathbf{s'}$).
While we can adopt this conceptually, we require an
efficient representation to enable \emph{dual}
interpretation in a \emph{single} algorithm pass.
To solve this problem, we encode each lattice node
(subset) as a \emph{mask}, where each mask element
reflects one of two possible states
(\keyword{True} or \keyword{False}) for the
input $s \in \mathbf{s'}$ it represents.
Masks permit a single canonical representation for each
lattice node while simultaneously permitting two
interpretations: one where
\keyword{True} indicates \emph{retention} of the
corresponding input (and omission when \keyword{False}),
and one where \keyword{True} indicates \emph{omission}
(and retention when \keyword{False}).

\textbf{Pruning Adjustments (Observation \#2):}
Although our Mono Rule Miner pruning strategy can still
be used to prune invalid rules, we can only prune rules
of the same type.
Therefore, even if no further rules of one type can be
found, lattice navigation will continue if rules of the
other type may still be found.
In short, Dual pruning opportunities can only be less
common (and thus less effective overall) than Mono,
since a subset must be invalid under \emph{both}
rule types before descendants can be pruned from the
search.
Moreover, once navigation reaches the bottom half of the
lattice, the concrete subset for a
given rule may have already been posed to the model
$M$ when interpreted under the opposite type.
In effect, this results in the
$\mathcal{O}({2^{|\mathbf{s}|}})$
time complexity being multiplied by a constant factor
of 2.

\begin{algorithm}[t]
\caption{Dual Rule Miner}
\label{algo:dual}
\begin{algorithmic}[1]

\Require The input set $\mathbf{s}$
\Require The remaining inputs $\mathbf{c}$
\Require The retention output predicate $O_{ret}$
\Require The omission output predicate $O_{omi}$
\Require The model $M$
\Ensure The rule-based explanations
\vspace{10pt}

\State $R_{ret} \gets \{\}$ \Comment{Initialize valid retention rules}
\State $R_{omi} \gets \{\}$ \Comment{Initialize valid omission rules}

\State $Z_{ret} \gets \{\}$ \Comment{Initialize invalid retention nodes}
\State $Z_{omi} \gets \{\}$ \Comment{Initialize invalid omission nodes}

\For{$l' \in [0, |\mathbf{s}|]$}
\label{linenum:dual:level_loop}

    \State $l \gets |\mathbf{s}| - l'$
    \Comment Compute current level size

    \State $S \gets \{ \mathbf{s'} \subseteq \mathbf{s} \mid |\mathbf{s}'| = l \}$
    \Comment Get all current level nodes

    \State $Z_{ret} \gets \{ \mathbf{z} \subset \mathbf{z'} \mid \mathbf{z'} \in Z_{ret} \land |\mathbf{z}| = l \}$
    \Comment{Get children of invalidated parents}
    \label{linenum:dual:level_invalid_parents_ret}

    \State $Z_{omi} \gets \{ \mathbf{z} \subset \mathbf{z'} \mid \mathbf{z'} \in Z_{omi} \land |\mathbf{z}| = l \}$
    \label{linenum:dual:level_invalid_parents_omi}

    \State $Q' \gets \{ (\mathbf{s'}, [\mathbf{s'} \notin Z_{ret}], [\mathbf{s'} \notin Z_{omi}]) \mid \mathbf{s'} \in S \}$
    \Comment Pair nodes with their validity
    \label{linenum:dual:level_validity_pairs}

    \State $Q \gets Q' \setminus \{ (\mathbf{s'}, v_{ret}, v_{omi}) \in Q' \mid \neg v_{ret} \land \neg v_{omi} \}$
    \Comment Prune totally invalid nodes
    \label{linenum:dual:level_prune_both_invalid}

    \For{$(\mathbf{s'}, v_{ret}, v_{omi}) \in Q$}
    
        \If {$v_{ret} \land O_{ret}( M(\mathbf{s'}, \mathbf{c}) ) = 1$}
        \Comment{Check for valid retention rule}
        \label{linenum:dual:predicate_check_ret}

            \State $R_{ret} \gets R_{ret} \cup \{ \mathbf{s'} \}$

        \Else
            \State $Z_{ret} \gets Z_{ret} \cup \{ \mathbf{s'} \}$
            \label{linenum:dual:record_invalid_ret}
        \EndIf

        \If {$v_{omi} \land O_{omi}( M(\mathbf{s} \setminus \mathbf{s'}, \mathbf{c}) ) = 1$}
        \Comment{Check for valid omission rule}
        \label{linenum:dual:predicate_check_omi}

            \State $R_{omi} \gets R_{omi} \cup \{ \mathbf{s'} \}$

        \Else
            \State $Z_{omi} \gets Z_{omi} \cup \{ \mathbf{s'} \}$
            \label{linenum:dual:record_invalid_omi}
        \EndIf

    \EndFor
    
\EndFor

\State \Return ($R_{ret}, R_{omi}$)

\end{algorithmic}
\end{algorithm}

\textbf{Example:}
The implementation of Dual Rule Miner is given by
Algorithm \ref{algo:dual}.
The algorithm requires the same inputs as
Mono Rule Miner, except that two separate predicates
$O_{ret}$ and $O_{omi}$ are expected for
retention and omission rule types, respectively.
Likewise, separate valid and invalid rule candidate
caches are now maintained for each rule type;
however, the familiar level-order traversal proceeds
in a similar fashion
(line \ref{linenum:dual:level_loop}).
Each subset of the current level size is evaluated
under both interpretations separately
(lines \ref{linenum:dual:predicate_check_ret} and
\ref{linenum:dual:predicate_check_omi}), each
applying their corresponding mask and being judged by
their corresponding output predicate.

The main point where Dual diverges from Mono is
the pruning step; i.e., just before looping
through a level's subsets.
Since a subset must be invalidated under \emph{both}
types before it can be pruned, Dual must reinitialize
the respective invalid candidate caches separately
(lines \ref{linenum:dual:level_invalid_parents_ret}
and \ref{linenum:dual:level_invalid_parents_omi}),
then prune only those subsets that do not
appear in \emph{either}
(lines \ref{linenum:dual:level_validity_pairs}
and \ref{linenum:dual:level_prune_both_invalid}).
For example, suppose that all subsets of size 4 and 3 have the
\emph{same} validity as indicated in
Figure \ref{fig:lattice} regardless of type
(i.e., $\{ s_1, s_2, s_3 \}$ is invalid for retention
and omission).
As the search continues to subsets of size 2, Dual
prunes the same children as described in
Section \ref{subsec:mono} for Mono, since these child
subsets are invalid under \emph{both} rule types.

Diverging from what is illustrated in
Figure \ref{fig:lattice}, suppose that $\{ s_1, s_4 \}$
is found to be a valid retention rule, but an invalid
omission rule.
Consequently, each descendant of this subset \emph{may}
be a valid retention rule, but can never be a valid
omission rule.
The algorithm inserts $\{ s_1, s_4 \}$ into the invalid
omission candidates cache
(line \ref{linenum:dual:record_invalid_omi}),
resulting in the inclusion of
its children $\{ s_1 \}$ and $\{ s_4 \}$
when reinitialized
(line \ref{linenum:dual:level_invalid_parents_omi}).
Since this subset and its children were not
included in \emph{both} invalid caches, the evaluation
loop will still ``evaluate'' these children.
However, per-type validity is still tracked and
associated with each evaluated subset
(line \ref{linenum:dual:level_prune_both_invalid}),
allowing the algorithm to ``short circuit'' and avoid unnecessary inference calls.

As with Mono Rule Miner, descendants of all invalid
rule candidates will \emph{continue} to propagate
through the continuous reinitialization of their
respective invalid candidate caches.
As a result, once a descendant of a subset invalidated
for one type is invalidated for the other type,
all descendants of this now ``totally invalidated''
descendant will be pruned.
Continuing our example, suppose that $\{ s_1 \}$
(already cached as an invalid omission rule via
propagation) is found to be an invalid retention rule
(via inference and judgment by $O_{ret}$).
This subset will be added to the invalid retention rule
candidates
(line \ref{linenum:dual:record_invalid_ret}),
and will therefore be present in both invalid caches,
meaning that all descendants (only $\{ \}$ in this case)
will be pruned by the algorithm.

\textbf{Analysis:}
The $\mathcal{O}({|\mathbf{s}| \choose l^*})$ space
complexity of Mono Rule Miner (established in
Section \ref{subsec:mono}) remains largely intact
here, but a constant multiple of 2 is again introduced
to account for each rule type requiring its own
rule validity cache.
In case the user wishes to avoid the aforementioned
repeated inference calls, Dual Rule Miner can cache
model responses, indexing them by their canonical mask.
Caching responses will raise the space complexity to 
$\mathcal{O}({2^{|\mathbf{s}|}})$.
We allow the user to decide whether they would prefer
repeated inference calls or higher memory use.
In Section \ref{subsubsec:pruning}, we empirically
study the proportion of overlapping inference calls in
a real RAG setting, and analyze the effectiveness of
Dual's caching in different conditions.

\section{Experimental Evaluation}
\label{sec:eval}

In this section, we conduct an experimental evaluation%
\footnote{%
    The code to reproduce both quantitative experiments is available at
    \url{https://github.com/joelrorseth/rag-llm-rule-miner}.
\label{footnote:eval}
}
to analyze the efficiency and quality of our rule mining
algorithms.
We begin with a qualitative evaluation that describes
a RAG case study in the domain of healthcare.
Then, we present a quantitative evaluation studying the
efficiency and scalability of our rule miners,
applied in both real-world and synthetic settings.
This is followed by another case study demonstrating
a large-scale application of our rule miners
in the domain of AI safety.
Lastly, we present another quantitative evaluation
studying the robustness of the rule miners and the
rules they find across different commercial LLMs.

\subsection{RAG Healthcare Case Study}
\label{subsec:casestudy}

\begin{figure*}[htbp]
\centerline{
    \includegraphics[width=1.0\textwidth]{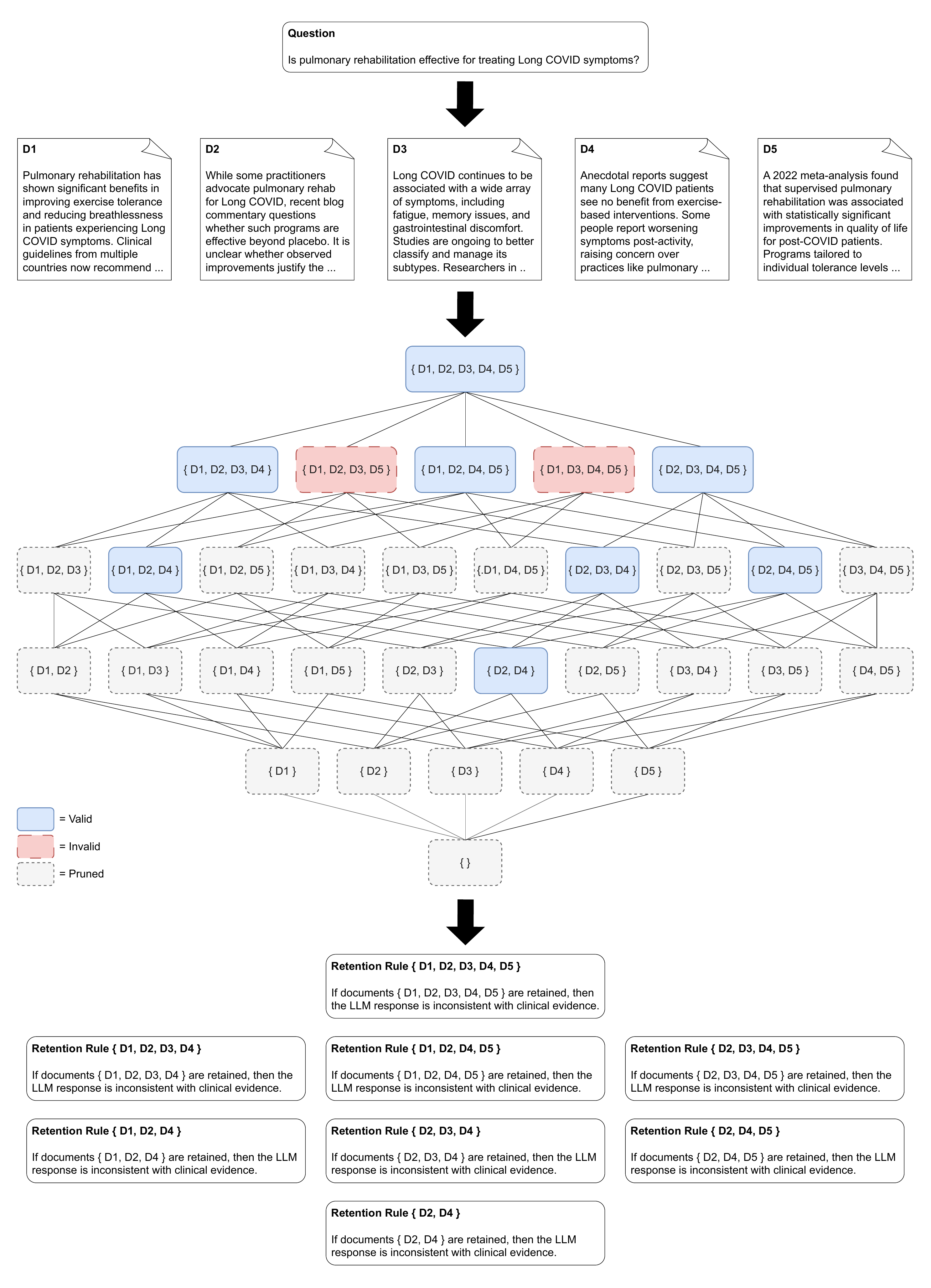}
}
\caption{
A lattice corresponding to the case study presented
in Section \ref{subsec:casestudy}.
Here, the minimal valid rule (lowest valid rule in
the lattice, as per Definition \ref{def:rule_minimality})
states that
``if D2 and D4 are retained, then the LLM produces a
response that is inconsistent with clinical evidence.''}
\label{fig:casestudy}
\end{figure*}

We now present a case study that considers
a real world scenario in the domain of healthcare.
We document several XAI challenges that arise when using
RAG-enabled LLMs in this setting, and analyze how
our rule-based explanations help to rectify them.
Our case study follows a technical user who is debugging
a medical research system used by healthcare
professionals (e.g., doctors, nurses) for the purposes
of real-time research.
One component of this system is a chatbot that enables
end users to pose domain questions and obtain answers
that are grounded in medical documents.
This chatbot component is implemented using a
retrieval-augmented LLM connected to a medical database.

The user is investigating a concerning trend
that several doctors have identified: when posed with
questions related to Long COVID, the chatbot often
returns a misinformed answer that downplays the
efficacy of pulmonary rehabilitation, despite clinical
consensus of its benefits.
The user suspects that one or more documents being
retrieved and provided to the LLM contain medical
misinformation, and that the LLM is ultimately grounding
its responses in this misinformation.
To dig deeper, the technical user wishes to understand
how the LLM's answer changes when seeded with different
documents provided via RAG.
Specifically, they wonder whether certain document
combinations lead the LLM to produce responses
inconsistent with the authoritative clinical evidence,
and therefore seek rules that link specific documents to
this inconsistency.

Suppose that the user is using a commercial
LLM augmented with a prompt-based RAG pipeline,
which works as follows.
Upon receiving a question from the end user, the
pipeline first prompts the LLM to generate one or more
keyword queries that can be posed to a retrieval model
to obtain documents that may prove helpful in
answering the question.
Then, for each generated keyword query, a proprietary
retriever capable of scoring document relevance
obtains the top-10 documents from the medical database
(i.e., top-10 deemed most relevant to the query).
The system establishes a final set of document sources
$\left\{ \text{D1}, \text{D2}, \text{D3}, \text{D4}, \text{D5} \right\}$
by sorting all retrieved documents by their relevance
scores and keeping only the 5 highest scoring.
These documents are inserted into a prompt that
instructs the LLM to use them to answer the
end user's question.

The user decides to debug a specific question,
``Is pulmonary rehabilitation effective for treating
Long COVID symptoms?,'' focusing first on the
documents returned by the system.
Upon review, the user confirms that the retriever
is functioning correctly, and that the retrieved
documents are relevant to the query.
The user suspects that, despite their relevance,
one or more of the retrieved documents may contain
conflicting or inaccurate information that causes the
LLM to produce a response inconsistent with clinical
evidence.
Therefore, the user seeks rules that will identify
these problematic documents and link them to the
inconsistent answers.

To generate the rules, the user defines an output
predicate that uses a secondary LLM to judge the
inconsistency of the response w.r.t. authoritative
clinical evidence (supplied by the user).
The user employs Mono Rule Miner to find all
valid retention rules.
As illustrated in Figure \ref{fig:casestudy},
several rules are returned, all
originating from a single minimal rule that states
the following:
whenever documents
$\left\{ \text{D2}, \text{D4} \right\}$ are retained,
the LLM response is inconsistent.
This suggests that, although neither D2 nor
D4 are sufficient individually to induce an
inconsistent answer (i.e., there is no valid rule
for $\left\{ \text{D2} \right\}$ or
$\left\{ \text{D4} \right\}$),
their \emph{combined} presence is.

With these new insights, the technical user can now
flag these two documents for further review by a
medical expert,
and temporarily remove one or both from the database
to prevent inconsistent answers.
After the expert revises the documents to ensure the content
is consistent with authoritative clinical evidence,
the same rule-based explanations can aid in
validating a successful redeployment. 
By running the rule miner again, the technical user
should find that no rules implicating inconsistent
answers exist, ensuring that the system is more likely to
remain faithful to the clinical consensus.

\subsection{Quantitative Evaluation: Rule Miner Efficiency}
\label{subsec:eval_efficiency}

In this section, we present a quantitative experiment that
evaluates the pruning improvements and scalability of our
rule miners, measured over both real-world and synthetic data.
We analyze rule miners' efficiency both absolutely and relatively,
discussing practical implications and directions for future improvement.

\subsubsection{Experimental Setup}
\label{subsubsec:eval_efficiency_setup}

To test our rule miners in a modern setting,
we design an experiment that tasks a RAG system with
question answering (QA), seeking rules to explain
its answer to specific questions.
We focus specifically on a RAG QA scenario where a factual
question is posed to an LLM along with a set of
knowledge sources for reference.
The LLM, $M$, is prompted to return the correct answer if
it can be inferred from the knowledge sources, or
``N/A'' otherwise.
The full implementation details of this ``LLM-under-test''
are provided in Appendix \ref{sec:appendix_prompts}.
To realize this experimental setting, we utilize the
\keyword{HotpotQA}\ dataset \citep{hotpotqa}, a QA benchmark
for LLMs and other QA systems.

\keyword{HotpotQA}\ tests a model's ability to answer
\emph{multi-hop} questions, which require consideration
of multiple independent pieces of information (``hops'')
to answer.
\keyword{HotpotQA}\ questions have exactly \textit{two} hops.
An example multi-hop question with two hops might ask,
``What is the chemical element named after
the physicist who developed the theory of relativity\@?''
Logically, answering this question requires one to
first learn that Albert Einstein primarily developed
the theory of relativity, then learn that the element
named after him is Einsteinium.
Each \keyword{HotpotQA}\ question is labeled with a
single ground truth answer, a set of
\textit{necessary} sources required to infer the answer
(constructed from \keyword{HotpotQA}\ ``supporting facts''),
and a set of
\textit{distractor} sources that are inconsequential
to the determination of the answer.
The sets of necessary and distractor sources are disjoint,
and there are exactly $k=2$ necessary sources for each question
(one for each hop).

As formalized in Section \ref{subsec:rule-formulation},
our rule miners require a definition for the input
set $\mathbf{s}$ and the output predicate(s) $O$.
To decide on a formulation, we must motivate our search
for rules.
We adopt an intuitive problem definition wherein a user
wishes to understand how the context sources provided
to the LLM (via \keyword{HotpotQA}) affect the correctness of the
LLM's answer.
Therefore, when posed with a specific \keyword{HotpotQA}\ question,
we define $\mathbf{s}$ as the set of sources that we
provide to the LLM.
The question and other input components constitute
$\mathbf{c}$, since they are not the focus of our
present formulation.

Meanwhile, we define two complementary output
predicates, $O_{consistent}$ and $O_{inconsistent}$,
which attempt to classify answers predicted by the LLM
based on their consistency w.r.t. the ground truth answer.
The predicate function first attempts to judge a predicted
answer's correctness via deterministic string matching
against the ground truth answer.
If the answers do not match, they are posed to an LLM
(\keyword{gpt-5-mini-2025-08-07}) that judges the equivalency
of the candidate answer w.r.t. the ground truth answer.
The full implementation details for the string matching
and ``LLM-as-a-judge'' procedures are provided in
Appendix \ref{sec:appendix_prompts}.
Under this formulation, our algorithms will mine
the following two types of rules:

\begin{enumerate}

\item \textbf{Retention rules:} Identify source
subsets that, when \textit{retained} within a
RAG prompt,
always induce an answer \textit{consistent} with
the ground truth answer.

\item \textbf{Omission rules:} Identify source
subsets that, when \textit{omitted} from a RAG prompt,
always induce an answer \textit{inconsistent} with
the ground truth answer.

\end{enumerate}

The goal of this quantitative evaluation is to measure
the efficiency of our rule miners, which
leverage novel pruning and early termination strategies,
as compared to a naive method that does not.
We propose that the efficiency of a rule miner can be
quantified by the average proportion of lattice nodes
that the rule miner actually visits.
In this setting, ``visiting'' a node implies executing a
single LLM inference call, therefore the number of calls
executed by the rule miner can also quantify its efficiency.
Expressing the number of nodes visited as a proportion allows
these measurements to be averaged and consolidated
over lattices of varying sizes (i.e., varying numbers of
possible nodes).

Using these two metrics, we proceed as follows.
We draw 50 QA pairs at random from the \keyword{HotpotQA}\
training set (\keyword{hotpot\_train\_v1.1.json}),
which will ultimately be posed to a single
``LLM-under-test'' $M$ (\keyword{gpt-5-mini-2025-08-07}).
For each question, we task each of our rule miners
(Mono seeking retention rules, Mono seeking omission rules,
and Dual seeking both) to
mine all rules arising from posing said question to $M$.
Given that $k=2$ (i.e., only two sources should be consequential
to consistency),
we anticipate more opportunity for pruning as additional distractor
sources are considered and $|\mathbf{s}|$ increases.
Therefore, we pose each question repeatedly to $M$ with increasingly
large $\mathbf{s}$ (spanning $1 \leq |\mathbf{s}| \leq 10$),
exhausting necessary sources before drawing additional
distractor sources.

To contrast the results of the real-world \keyword{HotpotQA}\
setting, we also compute results in a synthetic setting.
In this setting, we control for several real-world factors
that affect the true validity of rules found.
We create a dataset containing mock sources (both necessary and
distractor), then instantiate a mock LLM-as-a-judge and
LLM-under-test $M$ that use them to predict consistent vs.
inconsistent perfectly.
In effect, this controls for the non-determinism and stochasticity
of both LLMs, as well as for their propensity to make inaccurate
and inconsistent predictions.
We aggregate results over 1000 synthesized examples, and otherwise
perform the same steps as in the \keyword{HotpotQA}\ setting.
In the following, we discuss results and contrast them in the
\keyword{HotpotQA}\ vs. synthetic settings.

\subsubsection{Experimental Results: Pruning Improvements}
\label{subsubsec:pruning}

\begin{figure}
\centering
\begin{subfigure}[t]{.48\textwidth}
  \centering
  \includegraphics[width=1.0\textwidth]{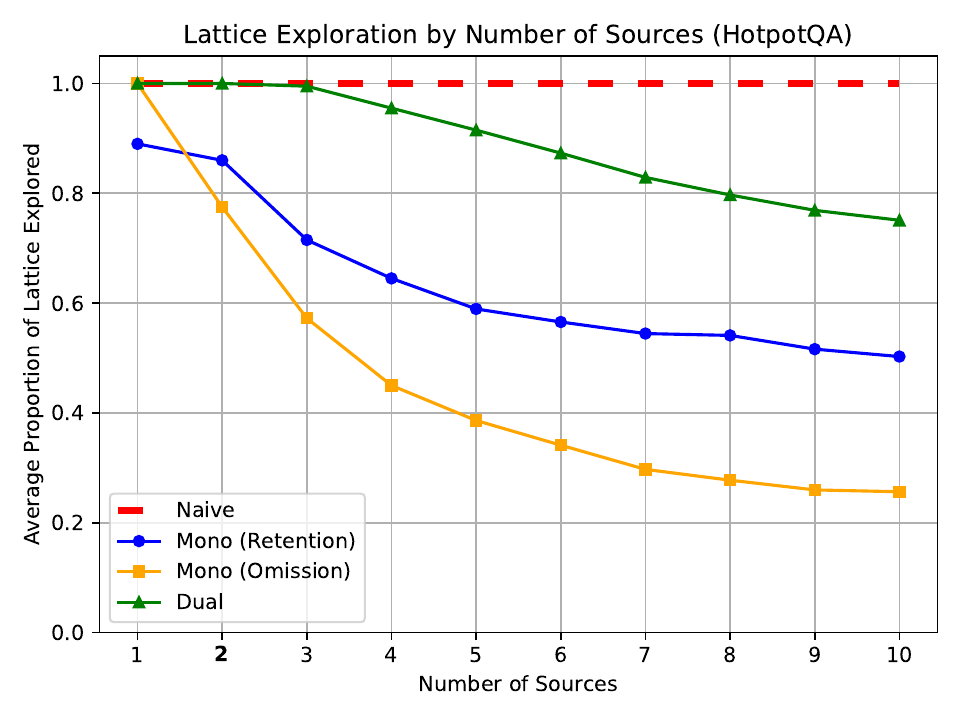}
  \caption{Rule miner lattice exploration as evaluated on a real
  LLM tasked with answering \keyword{HotpotQA}\ questions.}
  \label{fig:efficiency_exploration_hotpotqa}
\end{subfigure}%
\hspace{0.02\textwidth}
\begin{subfigure}[t]{.48\textwidth}
  \centering
  \includegraphics[width=1.0\textwidth]{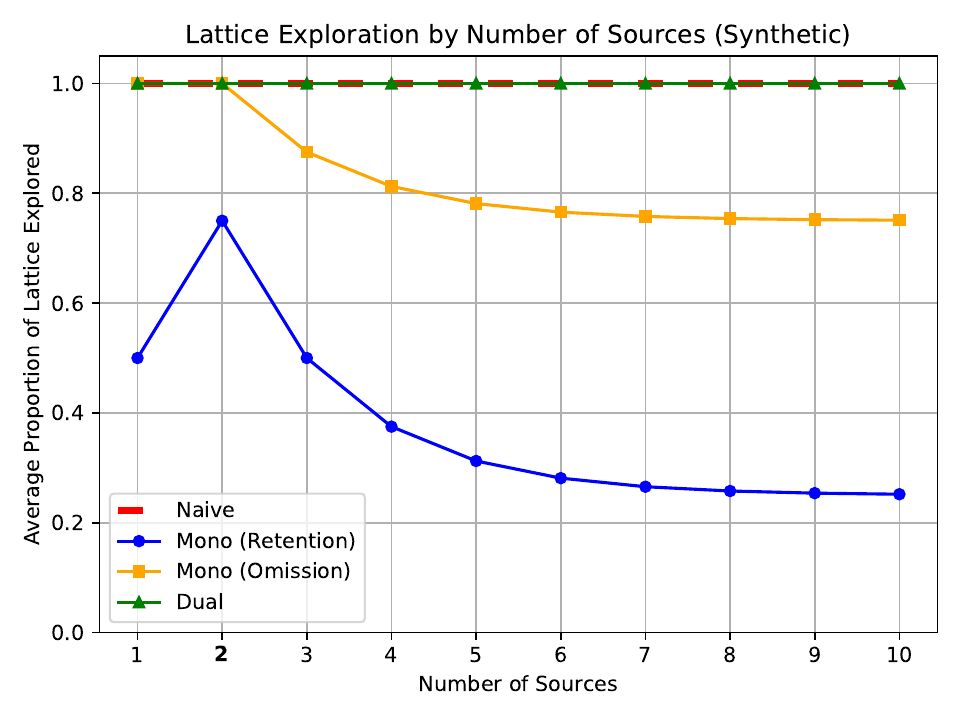}
  \caption{Rule miner lattice exploration as evaluated in a synthetic
  setting that controls for stochasticity and consistency.}
  \label{fig:efficiency_exploration_synthetic}
\end{subfigure}
\caption{The average degree of pruning achieved by different rule miners,
as measured by the average proportion of lattice nodes evaluated.
These measurements are plotted over increasing lattice sizes,
showing that our Mono and Dual rule miners prune more nodes
as more distractor sources are considered
(albeit at a slowing rate).}
\label{fig:efficiency_exploration}
\end{figure}

The results illustrated in Figure \ref{fig:efficiency_exploration}
demonstrate significant efficiency gains via pruning,
avoiding many of the $2^{|\mathbf{s}|}$ nodes that a naive algorithm
would evaluate.
Generally, once all \textit{necessary} sources are provided to
the LLM (i.e., $|\mathbf{s}| \geq 2$), the rule miners deem
additional sources inconsequential to the consistency of predicted
answers and prune larger portions of the lattice.
Figure \ref{fig:efficiency_exploration_hotpotqa} demonstrates that
Mono Rule Miner can prune as much as 50\% of the nodes when seeking
retention rules in the \keyword{HotpotQA}\ setting, and as much as
70\% when seeking omission rules.
The synthetic setting results in
Figure \ref{fig:efficiency_exploration_synthetic} reflect similar
trends with slightly smaller rates, indicating clear
asymptotic convergence at a 25\% pruning rate for omission rules
and 75\% for retention rules.
Across both settings, the proportion of this pruning increases with
$|\mathbf{s}|$, albeit at a slowing rate.

Mono Rule Miner pruning rates vary significantly by rule type,
but the trend flips in different settings:
more omission rules are pruned than retention rules in the
\keyword{HotpotQA}\ setting, but more retention rules
are pruned in the synthetic setting.
Naturally, the \textit{number} of valid rules found mirrors
this same pattern (see Figure \ref{fig:efficiency_rules}
in Appendix \ref{sec:appendix_rule_count}).
In our RAG QA formulation, retention rules must identify source
combinations sufficient to induce a consistent answer
(e.g., combinations with \textit{both} necessary sources),
but omission rules need only identify a combination whose
removal induces an inconsistent answer
(e.g., any combination containing \textit{one or more}
necessary sources).
Therefore, in a deterministic setting (like our synthetic
setting), we expect more omission rules than retention, but
the non-determinism introduced by using real LLMs with real
data (like our \keyword{HotpotQA}\ setting) defies these
expectations.
Due to several factors discussed in
Section \ref{subsubsec:robustness_llm_under_test},
especially the fact that the LLM guesses consistent answers
\textit{without} all necessary sources, fewer omission rules
hold than expected and more retention rules hold than expected.

\begin{figure}
\centering
\begin{subfigure}[t]{.48\textwidth}
  \centering
  \includegraphics[width=1.0\textwidth]{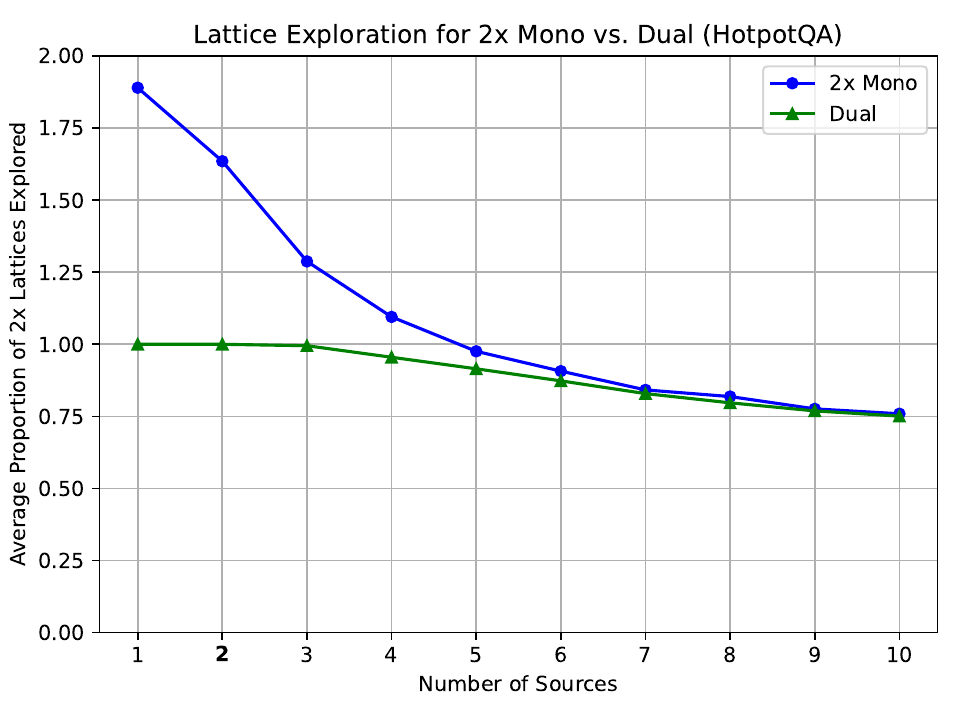}
  \caption{Mono vs. Dual Rule Miner lattice exploration as evaluated
  on a real LLM tasked with answering
  \keyword{HotpotQA}\ questions.}
  \label{fig:efficiency_mono_vs_dual_hotpotqa}
\end{subfigure}%
\hspace{0.02\textwidth}
\begin{subfigure}[t]{.48\textwidth}
  \centering
  \includegraphics[width=1.0\textwidth]{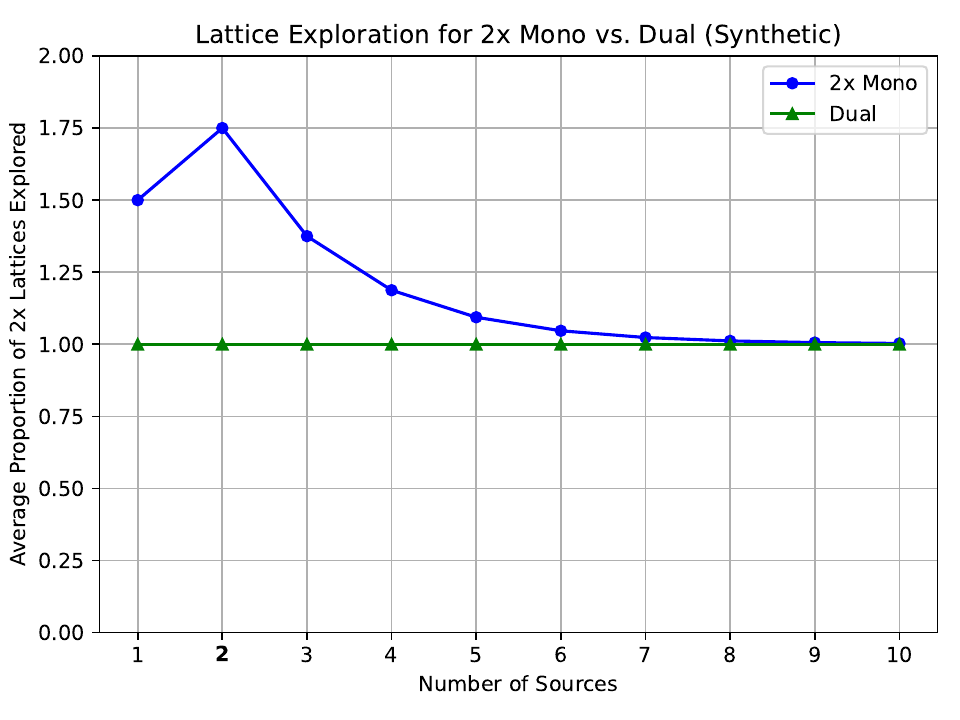}
  \caption{Mono vs. Dual Rule Miner lattice exploration as evaluated
  in a synthetic setting that controls for stochasticity and consistency.}
  \label{fig:efficiency_mono_vs_dual_synthetic}
\end{subfigure}
\caption{The average degree of pruning achieved by Dual Rule Miner
as compared to two equivalent Mono passes
(i.e., seeking retention rules, then omission).
Plotted over increasing lattice sizes and w.r.t. 200\% lattice
exploration, Dual avoids reevaluating a significant proportion
of lattice nodes, but to a lesser extent with additional
distractor sources.}
\label{fig:efficiency_mono_vs_dual}
\end{figure}

Figure \ref{fig:efficiency_exploration} reports that,
unsurprisingly, Dual Rule Miner evaluates most lattice nodes
and evaluates more nodes than either Mono run individually.
Importantly, it also confirms that Dual avoids evaluating any
nodes more than once, a key advantage over the alternative
strategy for finding both rule types: running Mono once for
each rule type.
Figure \ref{fig:efficiency_mono_vs_dual} provides a more
accurate juxtaposition of pruning efficiency for Mono vs. Dual,
rescaling ``exploration'' in terms of \textit{two} lattice passes.
Dual maintains an exploration rate near 100\% in both
\keyword{HotpotQA}\ and synthetic settings
(decreasing and eventually converging at 75\% in \keyword{HotpotQA}),
while Mono evaluates nearly 200\% (i.e., all nodes twice) in the worst
case.
As more distractor sources are considered, Mono's exploration
rate drops at a steady rate before eventually converging to
Dual's rate.
In short, Dual is more efficient for smaller sets of sources.

\subsubsection{Experimental Results: Scalability}
\label{subsubsec:scalability}

\begin{figure}
\centering
\begin{subfigure}[t]{.48\textwidth}
  \centering
  \includegraphics[width=1.0\textwidth]{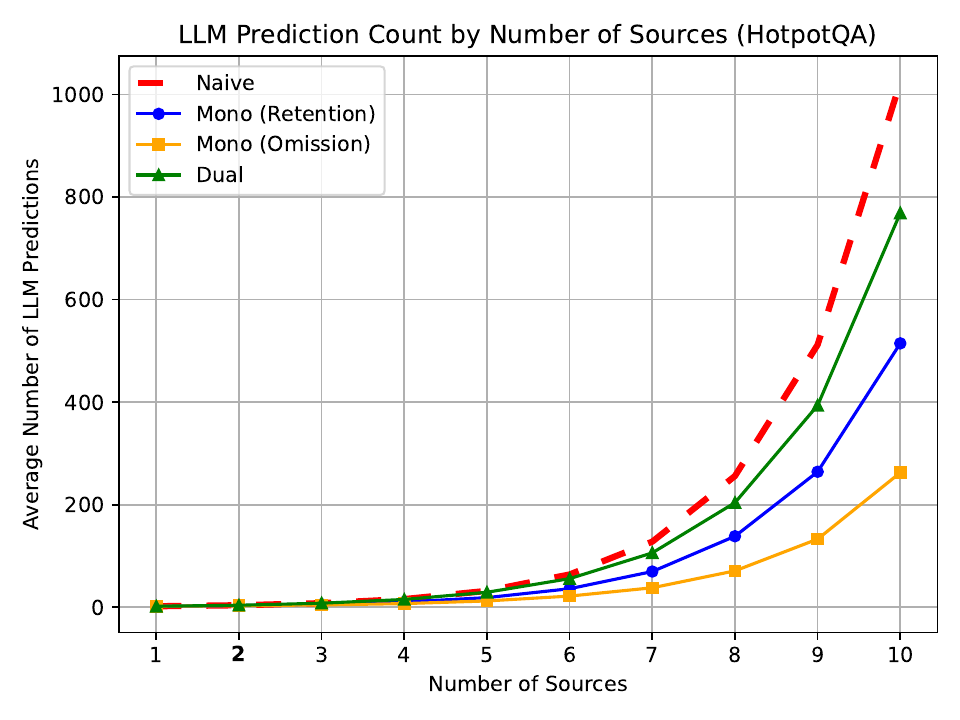}
  \caption{Rule miner predictions as evaluated on a real
  LLM tasked with answering \keyword{HotpotQA}\ questions.}
  \label{fig:efficiency_predictions_hotpotqa}
\end{subfigure}%
\hspace{0.02\textwidth}
\begin{subfigure}[t]{.48\textwidth}
  \centering
  \includegraphics[width=1.0\textwidth]{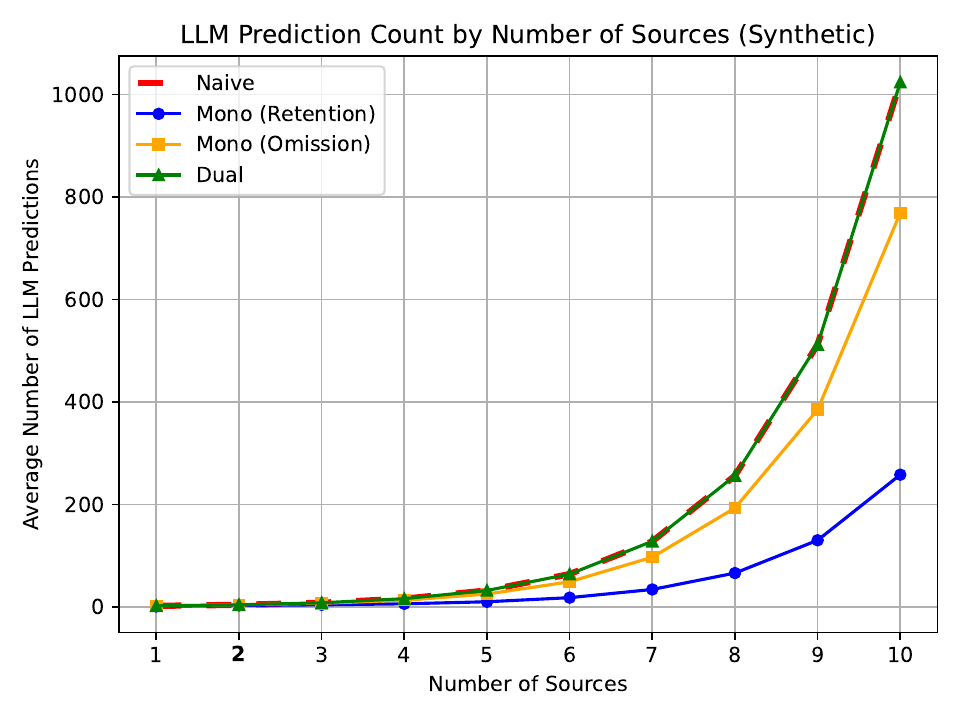}
  \caption{Rule miner predictions as evaluated in a synthetic
  setting that controls for stochasticity and consistency.}
  \label{fig:efficiency_predictions_synthetic}
\end{subfigure}
\caption{The average number of predictions made by different rule miners.
These measurements are plotted over increasing lattice sizes,
and are consistent with the exponential worst-case time complexity
of our rule miners.
Moreover, they establish $|\mathbf{s}| = 10$ as a tractable
upper bound for use of our rule miners with modern LLMs.}
\label{fig:efficiency_predictions}
\end{figure}

Interpreting the same data under a different lens,
Figure \ref{fig:efficiency_predictions} visualizes
and contrasts the scalability of each rule miner.
Specifically, it plots the number of predictions made by each
rule miner against the number of sources considered.
Generally, all rule miners improve upon the naive
algorithm that would evaluate \textit{all} $2^{|\mathbf{s}|}$
lattice nodes.
Moreover, the rate of scaling is consistent not only with the
worst-case time complexity derived in Section \ref{sec:algorithms},
but is highly consistent across both the \keyword{HotpotQA}\ and
synthetic settings.

By bounding this efficiency experiment to
$1 \leq |\mathbf{s}| \leq 10$, we also establish an
upper bound on the number of sources reasonably supported by
our rule miners at $|\mathbf{s}| = 10$.
We note that many real-world information retrieval (IR)
systems (e.g., web search engines) seek the top-10
documents most relevant to users' queries
\citep{JANSEN2006248, web-search-engines},
making this an appropriate boundary for explaining
real-world RAG LLMs that utilize such systems.
Figure \ref{fig:efficiency_predictions_hotpotqa} shows
that when $|\mathbf{s}| = 10$ in the
\keyword{HotpotQA}\ setting, Mono Rule Miner
finds all omission and retention rules using approximately
250 and 500 predictions respectively,
while Dual finds both using approximately 750.
Therefore, when $|\mathbf{s}|=10$, the prediction count of
Dual is similar to running Mono twice 
(minimal efficiency improvement),
but as $|\mathbf{s}|$ decreases,
Dual requires slightly less than double (moderate
efficiency improvement).
Given that many real-world RAG LLMs (e.g., LLMs tested in
these experiments) can complete QA inference calls with
sub-second response time, we assert that all rules can
be found in a few minutes when $|\mathbf{s}| \leq 10$.

When $|\mathbf{s}| > 10$, computing all rules can become
intractable due to the exponential time complexity.
This asymptotic boundary is well-studied in the association
rule mining literature, and is largely unavoidable due to
the exponential number of combinations that must be
evaluated in the worst case (the entire lattice).
The algorithms presented in this work adopt a design similar
to the original Apriori algorithm \citep{apriori}, which
ultimately navigates this exponential lattice
\textit{level-order}.
More efficient successors to Apriori have emerged in the
literature, and should inform successors of
the present work, but the exponential time complexity
remains.
We point interested readers to works such as
Eclat \citep{eclat}, whose depth-first traversal strategy
improves I/O overhead,
and FP-growth \citep{fpgrowth}, whose divide-and-conquer
strategy avoids candidate generation.

In the next subsection, we present a case study that describes
a method for tractably applying our rule miners when
$|\mathbf{s}| > 10$, so long as $k$ (the number of truly
necessary sources) is a small constant (e.g., $k \leq 10)$.
We contend that, although $k$ is not necessarily known upfront
by users of RAG systems, it is unlikely that both
$|\mathbf{s}|>10$ and $k>10$ in many practical settings.
IR research has shown that in real-world search engines,
the probability of a ranked document being relevant generally
decreases monotonically as rank increases
\citep{analysis_rank_dist}, suggesting that $k$ is likely
small in RAG settings.
Alternatively, rules can be mined for large $|\mathbf{s}|$ by
partitioning them (i.e., find rules separately for smaller batches)
or by merging them (i.e., find rules of coarser granularity).

\subsection{Scalable Red Teaming Case Study}
\label{subsec:scalability_case_study}

We now present a case study focused on a large-scale,
real-world scenario in the domain of AI safety.
This case study follows an AI safety researcher whose
team is developing advanced \textit{red teaming} methods
to safeguard a proprietary RAG LLM system.
These methods probe RAG LLMs adversarially in an attempt
to identify vulnerabilities, biases, and other harmful
behaviors.
The user is investigating a vulnerability recently
uncovered in the proprietary LLM: malicious instructions
can be injected into the RAG prompt through its retrieved
sources.
Given that users of the proprietary system are able to
supply their own RAG sources, patching this vulnerability
is imperative to ensure that the system's safety training
is not circumvented at inference time.

The user develops a novel solution that they believe will
catch prompt injections and refuse to respond.
To test the robustness of this solution, the team applies
their AI safety expertise to craft 50 adversarial RAG
sources with diverse injection strategies, all attempting
to induce a response containing a specific ``exploit hash''.
Though the user can quickly test each adversarial source
independently, they wish to test whether specific
\textit{combinations} of the adversarial sources are
effective in exploiting the vulnerability.
To efficiently identify any effective combinations,
the user seeks rules that associate specific combinations
with consistent, successful exploitation.

To generate rules, the user defines a simple output
predicate function that checks the output for the
exploit hash.
The user then provides all 50 adversarial sources
to Mono Rule Miner to find all retention rules,
but encounters a problem.
Given the large number of sources it is searching over,
the rule miner is proceeding at a very slow rate
and will never terminate.
Thinking of ways to reduce the dimensions of the problem,
the user realizes that since RAG ultimately concatenates
all sources into a single prompt, the 50 sources can also
be arbitrarily concatenated or grouped to reduce their count.
Once rules are mined for coarse-granularity grouped sources,
those not implicated by a rule can be discarded,
and finer granularity rules can then be mined by dividing
the remaining grouped sources again.

The user writes a script to automate these steps in a
divide-and-conquer fashion.
Beginning with 50 sources in one grouped source,
Mono Rule Miner confirms that the exploit is induced
(using 1 LLM inference call).
The grouping is therefore divided into 2 grouped sources
of size 25 and passed again to Mono, which finds a
minimal rule implicating both grouped sources jointly
(maximum of $2^2$ calls).
Since both grouped sources contain one or more sources
necessary to induce the exploit, the script continues by
dividing \textit{each} into two grouped sources of size
12 and 13, forming a new set of 4 grouped sources.
Mono finds a single rule implicating the first and fourth
grouped sources jointly (maximum of $2^4$ calls),
allowing for the second and third to be discarded.
The first and fourth are divided in two again and the
algorithm continues similarly; each iteration runs Mono
Rule Miner, discards grouped sources not implicated in minimal
rules, then divides remaining grouped sources and recurses
to solve them.

Once \textit{all} grouped sources become indivisible
(i.e., the original granularity is recovered),
the algorithm terminates and returns said sources.
After no more than 78 total LLM inference calls, the script
provides the user with a single minimal rule at the original
source-level granularity: retaining adversarial sources
$\left\{ \text{D8}, \text{D44} \right\}$ consistently
induces the exploit.
This suggests that $\text{D8}$ and $\text{D44}$ are jointly
but not independently effective in bypassing the user's
novel safety solution, perhaps due to interaction effects.
With these new insights, the user has been made aware of
the exploitative power of these sources when combined,
and can study them to refine their solution.

\subsection{Quantitative Evaluation: Rule Robustness}
\label{subsec:eval_robustness}

Building upon the real-world quantitative evaluation presented
in Section \ref{subsec:eval_efficiency},
we present additional quantitative experiments that
evaluate the robustness and quality of rules found by our
rule miners.
We first study the robustness of different ``LLMs-under-test'',
measuring the precision and recall of our rule miners in a
real-world \keyword{HotpotQA}\ setting.
Next, we study the robustness of the ``LLM-as-a-judge''
output predicate introduced in
Section \ref{subsubsec:eval_efficiency_setup},
using expert validation to estimate its precision and recall.

\subsubsection{Experimental Setup}
\label{subsubsec:eval_robustness_setup}

In this robustness evaluation, we follow and build
directly upon the efficiency evaluation setup
described in Section \ref{subsubsec:eval_efficiency_setup}.
The goal of the present evaluation is to measure the
robustness of our rule miners (which mine ``exact'' rules)
when used with real-world LLMs (whose non-determinism
can induce answer stochasticity).
We aim to measure and analyze the consistency of the
``LLM-under-test'' and ``LLM-as-a-judge'' components,
since their stochasticity and propensity for error can lead
to false positives and false negatives in their respective
functions.
Naturally, this evaluation focuses exclusively on the
real-world setting, as the synthetic setting (introduced
also in Section \ref{subsubsec:eval_efficiency_setup})
deliberately controls for the aforementioned components.
Therefore, we once again adopt \keyword{HotpotQA}\ to
conduct a similar multi-hop QA experiment.

Building upon the previous \keyword{HotpotQA}\ experiment,
we adopt the same definition for retention and omission rules,
as well as the same implementation for output predicates
$O_{consistent}$ and $O_{inconsistent}$.
However, the shift in experimental focus from efficiency to
robustness necessitates a different parameterization.
We propose that the robustness of the LLM components can
ultimately be quantified in terms of \textit{precision} and
\textit{recall} by comparing actual vs. expected rules.
After running a rule miner to obtain the ``actual'' rules,
precision measures how many actual were expected
(valid), while recall measures how many expected
rules were actually output.
We construct ground truth ``expected'' rules from the
\keyword{HotpotQA}\ QA pairs to facilitate this computation;
retention rules must retain both necessary sources to induce
a consistent answer, while omission rules must omit one or
more necessary sources to induce an inconsistent answer.

Using these metrics, we proceed similarly to the steps
described in Section \ref{subsubsec:eval_efficiency_setup}.
We draw 50 QA pairs from \keyword{HotpotQA}\ training set
at random, and fix each with exactly 5 sources: both
necessary sources and three additional distractor sources.
We pose these questions to three different commercial RAG LLMs,
each initialized with a \textit{temperature} of 1.0:
OpenAI's GPT-5 Mini (\keyword{gpt-5-mini-2025-08-07}),
Google's Gemini 2.5 Flash (\keyword{gemini-2.5-flash}),
and Anthropic's Claude Haiku 4.5 (\keyword{claude-haiku-4-5-20251001}).
This enables us to contrast robustness across several
commercial LLMs of similar capability.

We task Dual Rule Miner with finding all rules (of both types)
for each LLM posed with each question, but do so repeatedly to
obtain 10 \textit{sampled} LLM answers for each.
Due to their inherent stochasticity, LLMs may produce any number
of different outputs when posed repeatedly with the same
input prompt, potentially contradicting itself.
Many works in both the machine learning and LLM literature have
formalized inference strategies that embrace this reality
(e.g., ensemble methods \citep{ensemble_methods, llm_ensemble}),
often improving overall accuracy by sampling multiple answers and
computing a single final answer (e.g., via majority voting).
Building upon these works, we compute a majority answer
for each sample size in the range of 1 to 10,
assessing the sample group as consistent only when 50\%
or more individual samples are judged consistent
(see Appendix \ref{sec:appendix_prompts} for more detail).
We then contrast and analyze the evolution of robustness over
these increasing sample sizes, thereby measuring the
robustness of our ``exact'' rules in the face of
answer stochasticity.

\subsubsection{Experimental Results: ``LLM-under-test'' Robustness}
\label{subsubsec:robustness_llm_under_test}

\begin{figure}[t]
\centering
\begin{subfigure}{.48\textwidth}
  \centering
  \includegraphics[width=\textwidth]{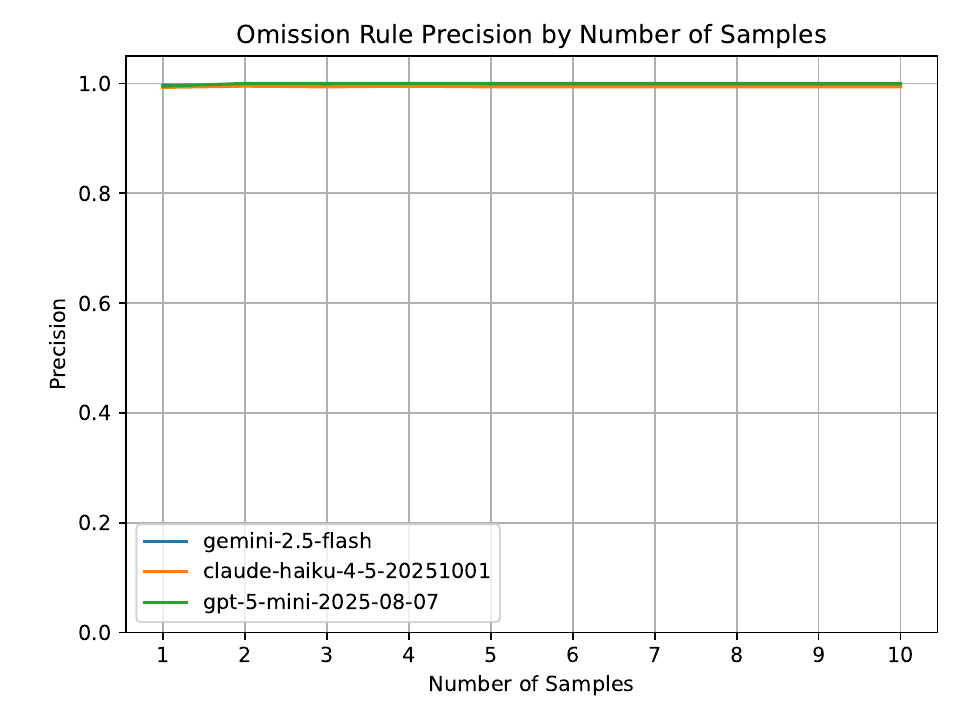}
  \caption{Precision of omission rules generated.}
  \label{fig:robustness_omission_precision}
\end{subfigure}
\hfill 
\begin{subfigure}{.48\textwidth}
  \centering
  \includegraphics[width=\textwidth]{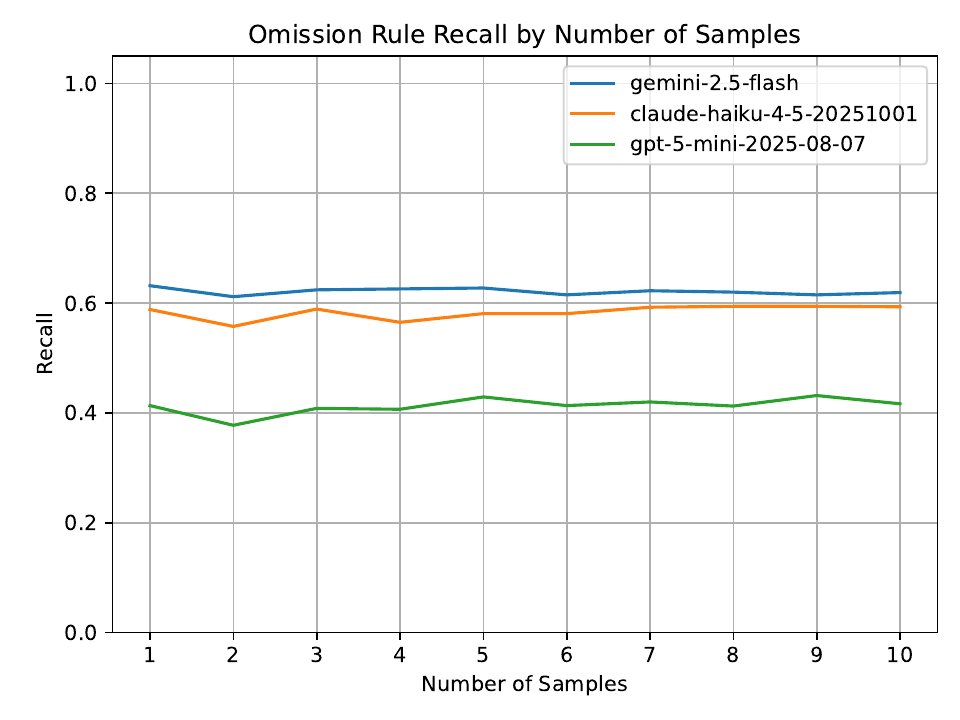}
  \caption{Recall of omission rules generated.}
  \label{fig:robustness_omission_recall}
\end{subfigure}


\begin{subfigure}{.48\textwidth}
  \centering
  \includegraphics[width=\textwidth]{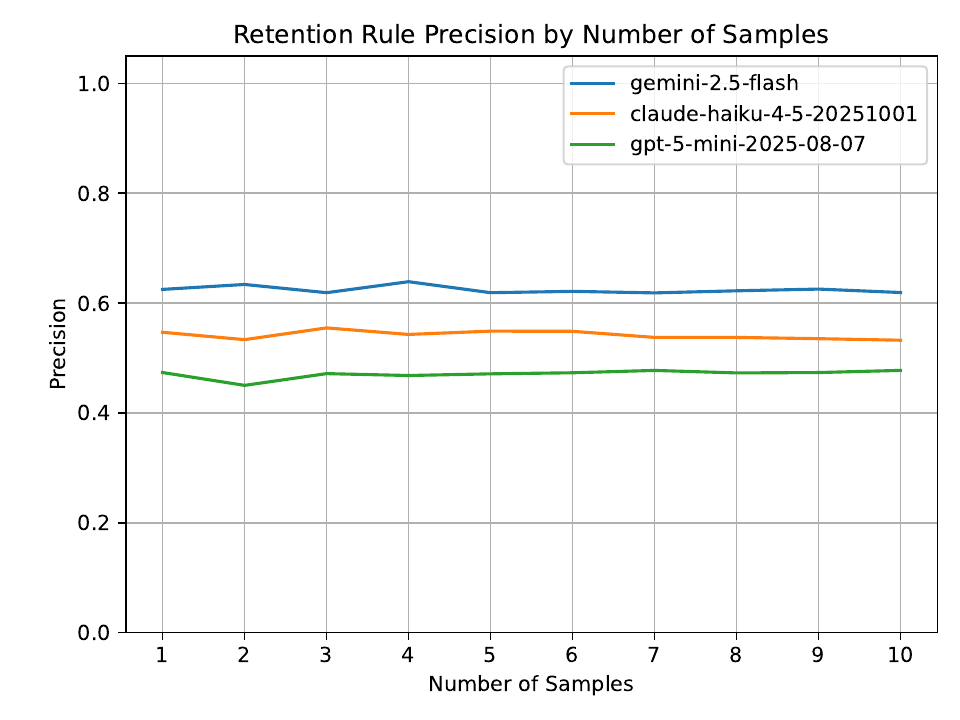}
  \caption{Precision of retention rules generated.}
  \label{fig:robustness_retention_precision}
\end{subfigure}
\hfill
\begin{subfigure}{.48\textwidth}
  \centering
  \includegraphics[width=\textwidth]{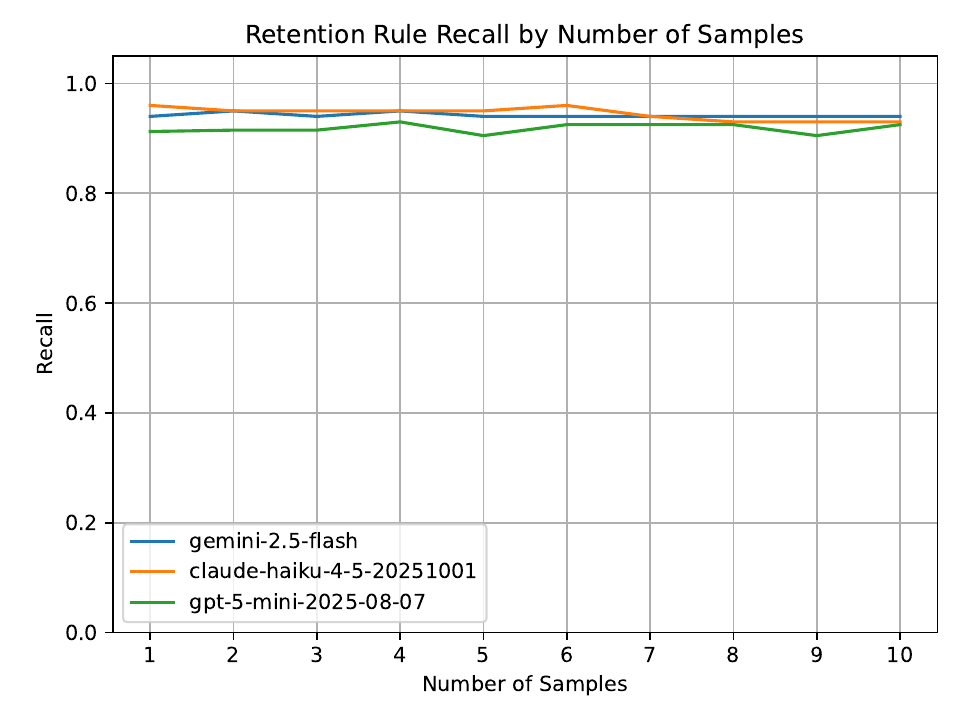}
  \caption{Recall of retention rules generated.}
  \label{fig:robustness_retention_recall}
\end{subfigure}
\caption{The precision and recall of rules generated using different
``LLMs-under-test'' over an increasing number of answer samples.}
\label{fig:robustness}
\end{figure}

In Figure \ref{fig:robustness}, we present the robustness of
the different ``LLMs-under-test'', reporting on four pairings:
precision and recall over both retention and omission rules.
In general, the chosen LLMs demonstrate similar robustness within
each pairing, and across all samples sizes therein.
With the exception of omission rule precision, for which all
LLMs score approximately 100\%, the results indicate that
Gemini 2.5 Flash is the highest performing, followed closely by
Claude Haiku 4.5 and then GPT-5 Mini.
Despite using a moderate LLM temperature of 1.0 to permit
stochasticity, all robustness measurements remain nearly
constant as more samples are considered for the majority vote.
This indicates that, despite the possibility of contradicting
each other, the stochastic outputs of LLMs can
(at least in similar QA settings) remain sufficiently
consistent to produce exact rules, regardless of sample size.

In Figures \ref{fig:robustness_omission_precision}
and \ref{fig:robustness_omission_recall},
we observe that omission rules exhibit near-perfect
precision but significantly lower recall.
This behavior reflects the fact that omission rules
are easier to satisfy; in theory, omitting a single
necessary source will suffice to induce an
inconsistent answer.
Therefore, more omission rules \textit{should} exist
than retention rules, as established in
Section \ref{subsubsec:pruning}.
It follows that the omission rules found are almost
always valid (high precision), since many exist.
Consequently, there are even more omission rules
that may not hold under a fallible LLM,
resulting in some being missed (lower precision)
in our ``exact rule'' formulation.

In Figures \ref{fig:robustness_retention_precision}
and \ref{fig:robustness_retention_recall},
we observe the opposite trend:
retention rules exhibit
near-perfect recall but significantly lower precision.
Although all necessary sources \textit{should} be
required to derive the ground truth answer, we
observe many instances where predicted answers are
consistent nonetheless.
We contend that this exceptional behavior stems from
data quality issues in the \keyword{HotpotQA}\ dataset,
as well as the LLMs' unexpected propensity to leverage
their pre-trained knowledge and ``guess''.
Consequently, many source combinations satisfy the
retention rule criteria \textit{in practice} despite
omitting one or both of the necessary sources (low precision).
Likewise, the presence of all necessary sources almost
always induces a consistent answer, leading to the
discovery of most expected retention rules (high recall).

Diving deeper to rationalize this unexpected behavior,
we note that several flaws have been documented in
\keyword{HotpotQA}\ questions, such as sources being
mislabeled as non-relevant
\citep{ni-etal-2021-mitigating, thakur-etal-2025-hard} and the
inadvertent introduction of reasoning shortcuts
\citep{jiang-bansal-2019-avoiding, chen-durrett-2019-understanding}.
Qualitatively analyzing our experimental results, we find
evidence that these flaws contributed to the
validation of unexpected retention rules (false positives) and
invalidation of expected omission rules (false negatives).

Specifically, we find that some distractor sources inadvertently contain
information relevant to the question, and can be used to induce 
a consistent final answer without all necessary sources.
For example, one question asks ``Which rock group was formed
earlier, Beastie Boys or The Shins?''
There are two necessary sources, one for each band; each states
the year the group was formed (1981 and 1996 respectively).
However, some distractors are unexpectedly relevant by virtue of
establishing pre-1996 release dates for some Beastie Boys songs.
So long as ``The Shins'' necessary source is retained,
a consistent answer can be inferred by also retaining any of
these ``release date'' distractors.

Furthermore, we note that general \keyword{HotpotQA}\
data quality issues and inconsistencies may also affect our results.
For instance, one question used in our experiments asks
``When was the composer who wrote the music for Euryanthe born?'',
but the ground truth answer ``31 January 179719'' is malformed.
In this same example, the first necessary source establishes that
``Carl Maria von Weber'' wrote the music for ``Euryanthe'',
but the second necessary source states only the malformed birth date
of the composer ``Franz Peter Schubert'',
who did not write ``Euryanthe''.
To illustrate another issue, we cite the question
``Which profession does Boris Barnet have in common with Baltasar Kormákur?'',
whose ground truth answer is ``film director''.
Despite the fact that \keyword{HotpotQA}\ questions should
have a unique answer, we find that the two necessary sources
indicate ``actor'' as another valid answer.

In cases where necessary information was truly absent,
the LLM often chose to defy its instructions and ``guess''
the answer instead of strictly inferring it.
Many of these guesses were consistent (and judged as such),
perhaps due to the fact that the commercial LLMs tested
likely trained on the same Wikipedia data from which
\keyword{HotpotQA}\ was constructed.
The remaining inconsistent guesses accounted for much of the
stochasticity observed across answer samples.
We note that answer diversity can be expected to increase as
temperature is increased, and decrease as temperature is decreased.
However, since diversity is offset by our ``majority answer''
sampling, we would not expect rule validity to be significantly
impacted if our experiments were repeated with a temperature of 0.

\subsubsection{Experimental Results: ``LLM-as-a-judge'' Robustness}
\label{subsubsec:robustness_llm_as_a_judge}

To assess the reliability of the ``LLM-as-a-judge''
used across all of our quantitative experiments,
we conduct an evaluation of its precision and recall
utilizing ground truth answers.
We randomly sample 100 judgments made by the
``LLM-as-a-judge'' during the robustness experiments
(inspecting all 10 answer samples),
drawn randomly across all QA pairs and
``LLMs-under-test''.
Each judgment is reviewed for
correctness by an expert validator, who assesses its
correctness by comparing the consistency of the predicted
answer w.r.t. the \keyword{HotpotQA}\ ground truth answer.
By tracking the number of examples that were accurately
and inaccurately judged as ``consistent'' and ``inconsistent'',
we can compute the precision and recall of the
``LLM-as-a-judge''.

Across the evaluated judgments, the ``LLM-as-a-judge''
achieves precision of 98.89\% and recall of 100\%,
indicating that nearly all answers judged ``consistent''
were correct, and \textit{all} truly ``consistent''
answers were correctly judged as such.
The false positive rate is 0.90\%, indicating that extremely
few ``inconsistent'' answers were judged ``consistent''
by mistake.
These results indicate that the judge is highly reliable at
identifying answers consistent with the ground truth answer,
and makes few mistakes in this QA setting.
Moreover, we observe that the ``LLM-as-a-judge'' judges
consistency at a semantic level, affording a level of nuance
that makes it robust to unexpected inconsistencies.

Specifically, we observe that the ``LLM-as-a-judge'' is robust
in identifying alternative answers that are consistent with
the ground truth answer provided.
For example, the \keyword{HotpotQA}\ question
``Kevin Olekaibe plays for the team that competes in what league?''
is labeled with the ground truth answer ``NBA G League''.
This league is known more generally as the NBA's developmental
league, and has been renamed multiple times (most recently to
``NBA G League'', named after its current sponsor ``Gatorade'').
This question introduces ambiguity, as previous names for
the league may or may not be consistent with ground truth.
Expert validation finds that the ``LLM-as-a-judge'' exhibits
semantic understanding of the answers in the context of the
question, and judges accordingly.
In one response to this question, the ``LLM-as-a-judge''
correctly inferred that ``NBA Development League'' is a more
general term to describe the ``NBA G League'', and correctly
judged it as ``consistent''.

Beyond robustness to diverse answers, we also observe robustness
to diverse LLM responses.
Our experiments show that Claude Haiku 4.5,
despite being limited to 100 output tokens and prompted to
output only the final answer, frequently outputs reasoning text
that exhausts the allotted output tokens and is
eventually truncated (i.e., before establishing a final answer).
In most cases, these responses should be judged as inconsistent
since a final answer was not clearly established, but reasoning
sufficiently conclusive and aligned with ground truth
could qualify as consistent.
We note that across truncated responses observed via expert
validation, those misjudged as ``consistent'' typically proposed
the ground truth answer as a candidate answer, but were truncated
shortly before articulating a final answer.
These situations demonstrate how judgment of reasoning-heavy or
partial answers is nuanced and potentially subjective, but
the ``LLM-as-a-judge'' excels in this setting.
Expert validation confirms that these truncated
responses are rarely misjudged, as evidenced by the low false
positive rate.

\section{Conclusions}\label{sec:conclusion}

We proposed the first formulation
of rule-based explanations for retrieval-augmented
LLM systems.
We presented two complementary variants that leverage
the concept of feature ablation to link certain
input features to the satisfaction of arbitrary
output predicates.
We implemented efficient algorithms for mining these
rules, which make use of pruning and early
termination.
By way of quantitative and qualitative evaluations,
we demonstrated the efficiency of our algorithms and
the utility of our explanations, as applied in
real-world RAG scenarios.

Our novel formulations and algorithms invite
promising directions for future work.
For instance, a relaxed version of our formulation
could model rules that hold \emph{approximately}
(i.e., with some minimum confidence) rather than
\emph{exactly} (i.e., with certainty), which could
help model the nuances of LLM non-determinism.
Furthermore, with our rule mining algorithms,
\keyword{HotpotQA}\ and similar benchmarks present an
opportunity for novel benchmarking studies.
Since expected rules can be derived from
\keyword{HotpotQA}'s ground truth answers and supporting
sources, a study could empirically compare LLMs on
the basis of whether rules expected to hold
actually do.
Similarly, this ground truth data could be used to
assess how often an LLM hallucinates answers not
present in retrieved sources, or how often it
deviates from RAG prompt instructions.

\backmatter








\section*{Declarations}

\subsection{Ethics Approval and Consent to Participate}
Not applicable.

\subsection{Consent for Publication}
Not applicable.

\subsection{Availability of Data and Material}
\label{subsec:material}

Implementations of our proposed algorithms, along with all
code and data used to conduct our quantitative experiments,
are available in a public repository hosted at
\url{https://github.com/joelrorseth/rag-llm-rule-miner}.
The synthetic data used in our quantitative experiments is generated
by the experiments at runtime in a fully reproducible manner.
The \keyword{HotpotQA}\ dataset \citep{hotpotqa} used in
our quantitative experiments has been made available publicly
by its authors, but instructions for downloading it are
included in our repository.

\subsection{Competing Interests}
The authors declare no competing interests.

\subsection{Funding}
This research was supported in part by the
Natural Sciences and Engineering Research Council of Canada (NSERC)
through a PGS D scholarship (awarded to the first author)
and Discovery Grant funding (awarded to multiple
co-authors).
One author was supported by the Connected Minds program,
funded by the Canada First Research Excellence Fund (CFREF).

\subsection{Author Contributions}
Conceptualization:
Joel Rorseth, Parke Godfrey, Lukasz Golab,
Divesh Srivastava and Jarek Szlichta.
Methodology, software, data curation, formal analysis,
investigation, visualization, and writing – original draft:
Joel Rorseth.
Supervision and writing – review and editing:
Parke Godfrey, Lukasz Golab,
Divesh Srivastava and Jarek Szlichta.
All authors read and approved the final manuscript.






\begin{appendices}

\section{LLM Prompt Templates for Experimental Evaluation}
\label{sec:appendix_prompts}

Each of the ``LLMs-under-test'' in the quantitative evaluations is
posed with a prompt constructed from the template in
Listing \ref{lst:qa_prompt}.
The text for the given question and knowledge sources is inserted into
the corresponding template variables, along with additional newlines to
delineate individual sources.
If no sources are given, prediction is avoided any ``N/A'' is returned.
For all experiments and LLMs, the temperature parameter is set to 1.0,
reasoning effort to the minimum allowed value, and maximum number of
output tokens to 100.
Before drawing \keyword{HotpotQA}\ questions for the experiments,
we discard those whose ground truth answer has more than 100 characters,
ensuring that the output token limit can be respected.

\begin{lstlisting}[caption={``LLM-under-test'' Prompt Template}, label={lst:qa_prompt}]
You are a question answering agent that uses knowledge sources to infer answers to questions.
You will be given a question and a set of knowledge sources.
Your goal is to determine the correct answer to the question.

**Question**
{question}

**Sources**
{sources}

**Task**
Determine the correct answer to the question using the knowledge sources.
If an answer cannot be inferred from the knowledge sources, return N/A.

**Output**
Return only the final answer, and nothing else.
\end{lstlisting}

Each answer generated by an ``LLM-under-test'' is judged w.r.t.
its respective ground truth answer,
first using string matching to identify near-exact matches,
then using ``LLM-as-a-judge'' if no match was found.
The string matching procedure checks for equality between a candidate
and ground truth answer after performing case folding,
punctuation stripping,
NFKD (Normalization Form Compatibility Decomposition),
and ASCII encoding to remove diacritics.
If the strings do not match, the candidate and ground truth answer
are posed to an LLM (along with the question) to classify
their semantic equivalence (``consistency'').
These pieces of information are substituted into the prompt template
in Listing \ref{lst:grader_prompt}, which is subsequently posed to
the LLM.

In the robustness experiments presented in
Section \ref{subsec:eval_robustness},
we pose the same prompt to each ``LLM-under-test'' up to
10 times.
We compute a final judgment for the sampled answers using
binary majority aggregation:
the ``final answer'' is judged ``consistent'' only if 50\% or
more of the sampled answers are judged ``consistent''.
Since sample sizes are tested in increasing order,
we cache previous responses for all ``LLMs-under-test''
to improve experimental efficiency, computing only a
single new answer sample in each round.
To further improve efficiency, all sampled answers
(for the same question) are \textit{batched} into a single
``LLM-as-a-judge'' prompt and assessed in one shot.
The prompt therefore instructs the LLM to assess the
equivalency of \textit{each} candidate (sampled answer)
against the ground truth answer.

\begin{lstlisting}[caption={``LLM-as-a-judge'' Prompt Template}, label={lst:grader_prompt}]
You are a question answering grading agent that assesses the equivalence of candidate answers against ground truth answers.
You will be given a question, ground truth that expresses the true answer, and a list of candidate answers.
Your goal is to assess whether each candidate is equivalent to the ground truth answer.

**Question**
{question}

**Ground Truth**
{ground_truth_answer}

**Candidate Answers**
{candidate_answers}

**Task**
Determine whether each candidate answer is equivalent to the ground truth answer, with respect to the given question.
Note that while the ground truth may include additional details, the candidate need only be equivalent to the final answer expressed therein.

**Output**
Return a list of booleans, one for each candidate answer in the same order they were given.
The boolean should be True if the corresponding candidate answer is equivalent to the ground truth answer, and False otherwise.
\end{lstlisting}

\section{Rule Count for Efficiency Evaluation}
\label{sec:appendix_rule_count}

\begin{figure}
\centering
\begin{subfigure}[t]{.48\textwidth}
  \centering
  \includegraphics[width=1.0\textwidth]{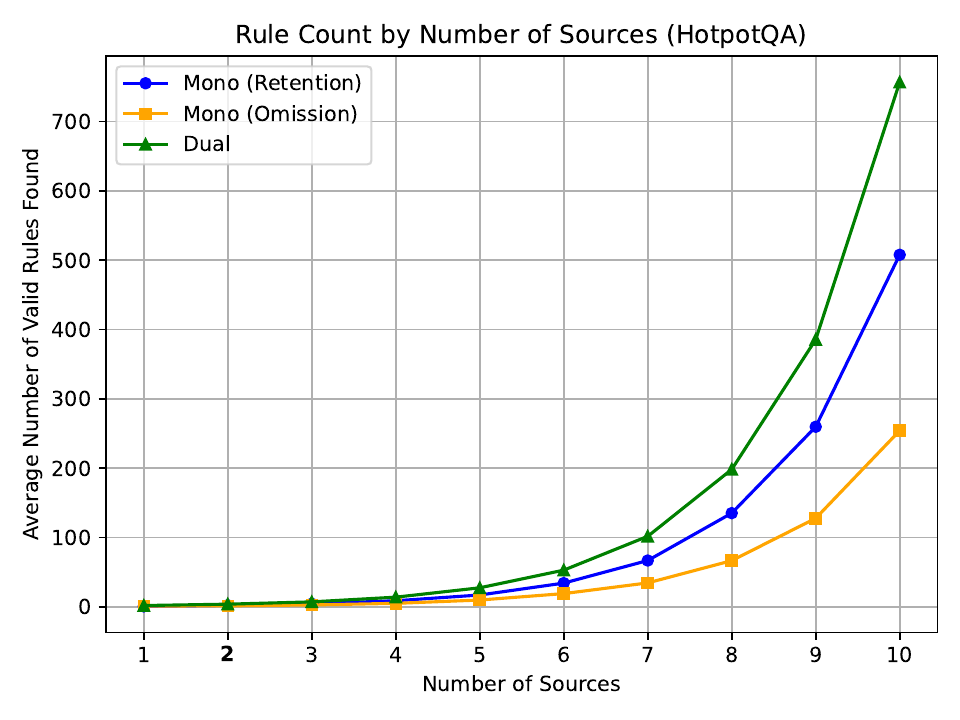}
  \caption{Rule count as evaluated on a real
  LLM tasked with answering \keyword{HotpotQA}\ questions.}
  \label{fig:efficiency_rules_hotpotqa}
\end{subfigure}%
\hspace{0.02\textwidth}
\begin{subfigure}[t]{.48\textwidth}
  \centering
  \includegraphics[width=1.0\textwidth]{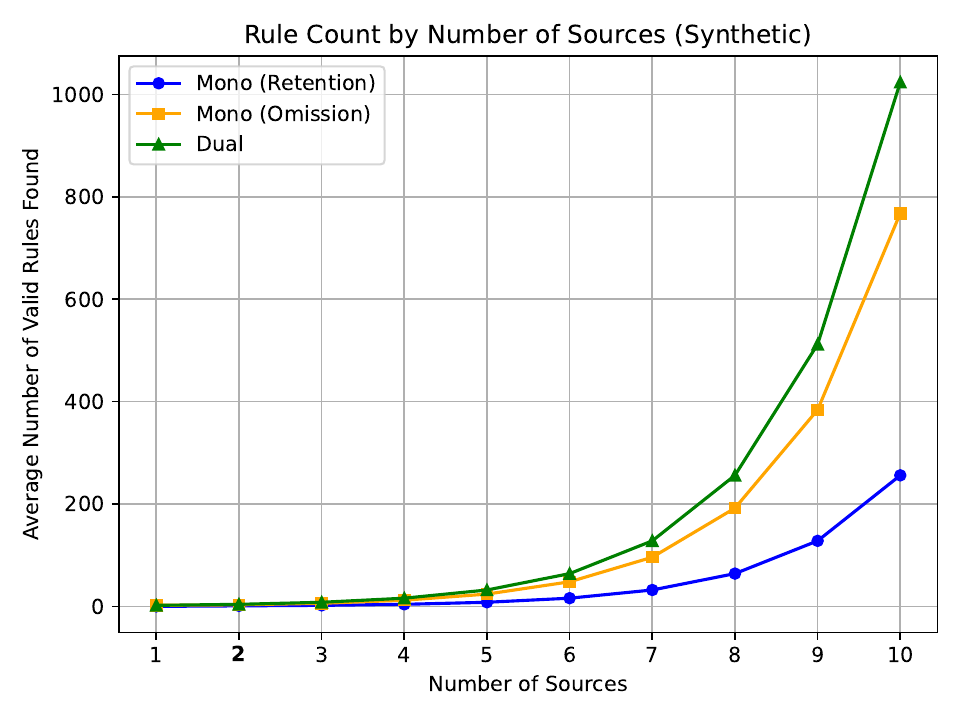}
  \caption{Rule count as evaluated in a synthetic
  setting that controls for stochasticity and consistency.}
  \label{fig:efficiency_rules_synthetic}
\end{subfigure}
\caption{The average number of valid rules found by different rule miners.
These measurements are plotted over increasing lattice sizes,
showing that our Mono and Dual rule miners find more rules as
more sources are considered.}
\label{fig:efficiency_rules}
\end{figure}

In Figure \ref{fig:efficiency_rules}, we illustrate the
number of rules found by each rule miner in our quantitative
efficiency experiment in Section \ref{subsec:eval_efficiency}.
The difference in \keyword{HotpotQA}\ vs. synthetic trends,
discussed in  Section \ref{subsec:eval_efficiency}, are evident
here as well.
Under our RAG QA formulation, more omission rules \textit{should}
exist than retention rules (as indicated in
Figure \ref{fig:efficiency_rules_synthetic}) since they are
easier to satisfy.
However, in the practical \keyword{HotpotQA}\ setting using
a commercial LLM, more retention rules are valid than
expected and fewer omission rules hold
(as indicated in 
Figure \ref{fig:efficiency_rules_hotpotqa}).
The trends are asymptotically proportional to the scalability
plots in Figure \ref{fig:efficiency_predictions}, which is why
they appear in the Appendix.




\end{appendices}

\bibliography{sn-bibliography}

\end{document}